\documentclass[journal=jcisd8,manuscript=article,layout=twocolumn]{achemso} 

\usepackage{chemformula} 
\usepackage[T1]{fontenc} 

\usepackage{url}
\usepackage{float}
\usepackage{caption}
\usepackage{amsmath}
\usepackage{graphicx}
\usepackage{listings}
\usepackage{float}
\usepackage{tabularx}
\usepackage{amsfonts}
\usepackage{hyperref}
\usepackage{listings}
\usepackage{verbatim}
\usepackage{xcolor}
\usepackage{soul}
\usepackage{fancyvrb}
\usepackage{mdframed}
\usepackage{adjustbox}
\usepackage{makecell}
\usepackage{comment}
\usepackage{enumitem}
\usepackage{booktabs}
\usepackage{arydshln}
\usepackage[final,markup=underlined,commandnameprefix=always]{changes} 

\definecolor{codebg}{RGB}{240,240,240}
\definecolor{codegreen}{RGB}{0,128,0}
\definecolor{codeblue}{RGB}{0,0,128}
\definecolor{codepurple}{RGB}{128,0,128}
\definecolor{codegray}{RGB}{128,128,128}

\lstdefinestyle{mystyle}{
    backgroundcolor=\color{codebg},
    commentstyle=\color{codegreen},
    keywordstyle=\color{codeblue},
    numberstyle=\tiny\color{codegray},
    stringstyle=\color{codepurple},
    basicstyle=\ttfamily\small, 
    breakatwhitespace=false,
    breaklines=true,
    captionpos=b,
    keepspaces=true,
    numbers=left,
    numbersep=5pt,
    showspaces=false,
    showstringspaces=false,
    showtabs=false,
    tabsize=2
}

\newcommand{\myverbatim}[1]{%
  \colorbox{codebg}{%
    \begin{minipage}{\dimexpr\linewidth-2\fboxsep}%
      \begin{verbatim}
      #1
      \end{verbatim}
    \end{minipage}%
  }%
}

\usepackage{caption}

\author{Morgan Thomas}
\affiliation[upf]
{Computational Science Laboratory, Universitat Pompeu Fabra, Barcelona Biomedical Research Park (PRBB), C Dr. Aiguader 88, 08003 Barcelona, Spain}
\alsoaffiliation[khalifa]
{Khalifa University of Science and Technology, Abu Dhabi, UAE}
\email{morganthomas263@gmail.com}

\author{Albert Bou}
\affiliation[upf]
{Computational Science Laboratory, Universitat Pompeu Fabra, Barcelona Biomedical Research Park (PRBB), C Dr. Aiguader 88, 08003 Barcelona, Spain}

\author{Jose Carlos Gómez-Tamayo}
\affiliation[JN]{In Silico Discovery, Janssen Research \& Development, Janssen Pharmaceutica N. V., Turnhoutseweg 30, B-2340 Beerse, Belgium}

\author{Gary Tresadern}
\affiliation[JN]{In Silico Discovery, Janssen Research \& Development, Janssen Pharmaceutica N. V., Turnhoutseweg 30, B-2340 Beerse, Belgium}

\author{Mazen Ahmad}
\affiliation[JN]{In Silico Discovery, Janssen Research \& Development, Janssen Pharmaceutica N. V., Turnhoutseweg 30, B-2340 Beerse, Belgium}

\author{Gianni De Fabritiis}
\affiliation[icrea]{Instituci\'o Catalana de Recerca i Estudis Avan\c{c}ats (ICREA), Passeig Lluis Companys 23, 08010 Barcelona, Spain}
\alsoaffiliation[upf]
{Computational Science Laboratory, Universitat Pompeu Fabra, Barcelona Biomedical Research Park (PRBB), C Dr. Aiguader 88, 08003 Barcelona, Spain}
\alsoaffiliation[acellera]{Acellera Labs, C Dr Trueta 183, 08005, Barcelona, Spain}
\email{g.defabritiis@gmail.com}

\title[An \textsf{achemso} demo]{REINFORCE-ING Chemical Language Models for Drug Discovery}

\begin{document}



\begin{abstract}
Chemical language models, combined with reinforcement learning (RL), have shown significant promise to efficiently traverse large chemical spaces for drug discovery. However, the performance of various RL algorithms and their best practices for practical drug discovery are still unclear. Here, starting from the principles of the REINFORCE algorithm, we investigate the effect of different components from RL theory including experience replay, hill-climbing, baselines to reduce variance, and alternative reward shaping. We propose a new regularization method more aligned to REINFORCE than current standard practices, and demonstrate how RL hyperparameters can be fine-tuned for effectiveness and efficiency. Lastly, we apply our learnings to practical drug discovery by demonstrating enhanced learning efficiency on frontier binding affinity models by using Boltz2 as a reward model. We share our RL models used in the ACEGEN repository, and hope the experiments here act as a guide to researchers applying RL to chemical language models for drug discovery.
\end{abstract}

\section{Introduction}\label{introduction}

Chemical language models (CLMs) \cite{grisoni2023chemical} are a widely-used \citep{martinelli2022generative} and effective approach to de novo molecule generation \citep{segler2018rnn, brown2019guacamol, polykovskiy2020molecular}. These models utilize molecular string representations to sequentially encode molecules one token at a time. In particular, the human-interpretable SMILES molecular grammar \citep{weininger1988smiles} performs consistently well \citep{skinnider2021chemical, skinnider2024invalid} despite the advent of new machine-learning-inspired grammars \citep{weininger1988smiles, o2023deepsmiles, krenn2020selfies}. Furthermore, the sequential nature of token-by-token molecule generation can be framed as a decision-making problem in which reinforcement learning (RL) can be applied.

The combination of CLMs with RL \citep{sutton2018reinforcement} is an evidenced strategy for a systematic exploration of chemical space \citep{olivecrona2017reinvent, popova2018reinforce, thomas2022augmented}. This approach leverages feedback from reward functions that assign a numerical value to molecular properties according to their desirability. Therefore, RL can progressively generate molecules that align better with predefined objectives, such as estimated potency, selectivity, bioavailability, or toxicity.


In recent work \cite{bou2024acegen}, we showed that REINFORCE-based \citep{williams1992simple} algorithms remain the gold standard when considering different challenges associated with molecule generation for drug discovery. This also aligns with recent studies \citep{ahmadian2024back} showing that when RL policies are pre-trained, as in the case of large and chemical language models, REINFORCE can outperform other algorithms generally considered more advanced, such as Advantage Actor Critic (A2C) \citep{mnih2016asynchronous} and Proximal Policy Optimization (PPO) \citep{schulman2017proximal}. We also found that policy regularization via reward shaping or additional loss terms that encouraged the agent to remain close to a prior was more effective than the explicit addition of constraints to the reward functions for maintaining favorable chemistry in practice. Furthermore, we deconstructed REINVENT into its independent components: the REINFORCE algorithm, experience replay, reward shaping, and a likelihood regularization loss term. We discovered that while the reward shaping is effective for regularization, it is not easily interpretable, preventing fine-grained control of the trade-off between exploration and exploitation. Additionally, we found that the likelihood regularization term does not improve the performance of the REINFORCE algorithm. 

In this paper, we test the effect of different orthogonal components inspired by RL literature on the REINFORCE algorithm. These include: the use of baselines to reduce variance in the gradient estimates, the impact of hill climbing by selecting the top-k elements in each data batch, different experience replay (ER) configurations, and a new more intuitive reward shaping that decouples the reward gradient from prior regularization. Lastly, we apply our learnings to a drug discovery challenge by optimizing frontier binding affinity models \cite{passaro2025boltz} to identify putative allosteric JNK3 ligands.

\section{Methods}
\subsection{Chemical language models combined with reinforcement learning}
CLMs \citep{grisoni2023chemical} that utilize language representation of chemistry such as SMILES \citep{weininger1988smiles} or DeepSMILES \citep{o2023deepsmiles} have shown promising success in de novo drug discovery \citep{olivecrona2017reinvent, grisoni2021combining, atz2024prospective, thomas2024modern}. Typical autoregressive models are trained unsupervised on a corpus of molecules at the task of ``next token prediction'', where all tokens combined sequentially form the full language representation of a molecule \citep{segler2018rnn}. A trained model can sample new tokens conditional upon previous tokens, such that the resulting molecules are likely to adhere to chemical rules and belong to a similar chemical distribution as the training molecules \citep{segler2018rnn, polykovskiy2020molecular}. 

From an RL perspective, the problem of sequentially building molecules using tokens can be viewed as navigating a partially observable Markov Decision Process (MDP) \citep{sutton2018reinforcement}, described by the quintuple $\langle S, A, R, P, \rho_0 \rangle$. Here, $S$ represents the set of all possible states in the problem space, and $A$ denotes the set of valid actions available to the agent. The reward function, $R : S \times A \times S \to \mathbb{R}$, assigns numerical values to transitions from one state to another based on the action taken. The transition probability function, $P : S \times A \to \mathcal{P}(S)$, specifies $P(s_{t+1}|s,a)$ as the probability of transitioning to state $s_{t+1}$ from the current state $s$ under action $a$. Lastly, $\rho_0$ signifies the initial state distribution.
In this context, a parameterized RL policy (a pre-trained CLM), denoted as $\pi_{\theta}$, guides molecule building by successively processing the current state, which represents a partially built molecule, and selecting actions $a_t$ that represent the next token at each step, until reaching a terminal state. The reward function evaluates the molecule with respect to a defined objective/s, assigning a scalar value either at each generative step or at the end of the process when a complete compound is produced. Within this framework, $\pi_{\theta}(a_t|s_t)$ denotes the probability of the policy function, parameterized by $\theta$, taking action $a_t$ in state $s_t$.
The sequence $\tau$ represents a complete series of actions needed to construct a molecule, often referred to as an episode. $P(\tau|\theta)$ indicates the probability of a trajectory $\tau$ given the policy parameters $\theta$. Additionally, $R(\tau)$ represents the cumulative sum of rewards over the trajectory $\tau$.

\subsection{Reinforcement learning with REINFORCE}

REINFORCE \cite{williams1992simple} is a policy gradient RL algorithm that seeks to learn policy parameters $\theta$ based on the gradient of a scalar performance measure $J(\theta)$, aiming to maximize performance as shown in \autoref{eq:reinforce}. Compared to algorithms like Proximal Policy Optimization (PPO), which are generally considered more advanced, REINFORCE offers lower computational costs and is less sensitive to hyperparameter tuning. PPO emphasizes stability through small, stable updates and is suited for scenarios with large off-policy gradient updates, a regime that dominates traditional deep RL benchmarks. However, this is not necessarily the case with pre-trained CLMs, where the policy weight initialization is not random due to unsupervised pre-training on a corpus of molecules constituting a prior policy. Additionally, in our setting, rewards are exclusively assigned to full generations, lacking true rewards for intermediary actions (or tokens in CLMs). Both of these characteristics align better with REINFORCE, which allows larger gradient updates and treats the entire generation as a single action.
This hypothesis was validated in the context of RLHF (RL from Human Feedback) \citep{ahmadian2024back}, and contributes to understanding why state-of-the-art methods for language-based drug discovery are based on REINFORCE.

\begin{equation}\label{eq:reinforce}
\nabla J(\theta) = \mathbb{E}_{\tau \sim \pi_{\theta}} \left[ \sum_{t=0}^{T} \nabla_{\theta} \log \pi_{\theta}(a_t|s_t) \cdot R(\tau) \right]
\end{equation}

\subsection{Extensions to REINFORCE}

By itself, REINFORCE has demonstrated effectiveness for language based drug discovery, as shown in our previous work \citep{bou2024acegen}. However, to further enhance its performance for the problem at hand, several orthogonal extensions of this algorithm have been applied in the literature to better optimize the reward function and generate chemically diverse, property-optimized molecules \citep{ahmadian2024back, thomas2022augmented, gao2022sample, olivecrona2017reinvent, blaschke2020reinvent, burda2018exploration, svensson2024diversity}. We re-evaluate a selection of extensions, aiming to combine them to create an integrated agent.

\textbf{Baselines:}
To enhance learning, the variance of the estimator in \autoref{eq:reinforce} can be reduced by subtracting any baseline 
\( b \) \chadded{as in }\autoref{eq:reinforce_baselines} which only depends on the state \citep{sutton2018reinforcement}. We consider two options: a moving-average baseline (MAB) and a leave-one-out baseline (LOO) \citep{ahmadian2024back}.

\begin{equation}\label{eq:reinforce_baselines}
\resizebox{0.9\columnwidth}{!}{$
\nabla J(\theta) = \mathbb{E}_{\tau \sim \pi_{\theta}} \left[ \sum_{t=0}^{T} \nabla_{\theta} \log \pi_{\theta}(a_t|s_t) \cdot (R(\tau) - b) \right]
$}
\end{equation}



\textbf{Hill-climb:}
Hill-Climb (HC) is a strategy where only a specific ratio of molecules generated by the agent is retained at each iteration and used on the training process. These selected molecules are chosen based on their reward ranking, with only the top-k performing molecules being kept. This method has been demonstrated to enhance learning efficiency in language-based de novo molecule generation \citep{thomas2022augmented}. In our study, we delve further into this approach by investigating the effects of retaining various portions of the data batch.

\textbf{Experience replay:}
While REINFORCE relies solely on on-policy data to refine its policy, some works have explored augmenting its learning process with off-policy data. This strategy involves storing the molecules with the higher rewards encountered during training, then randomly sampling batches of these off-policy data points. These batches are subsequently combined with the on-policy data batches generated by the agent during training. This approach has demonstrated enhanced sample efficiency in algorithms such as REINVENT \citep{gao2022sample} and PPOD \citep{libardi2021guided}. Notably, previous works primarily focused on sampling off-policy molecules with priority proportional to their reward. However, our study also explores a uniform sampling strategy.

\textbf{Policy regularization via reward shaping:}\label{sec:reward_shaping}
Reward shaping is the process of augmenting the rewards of RL agents to guide them towards desired behaviours that might not be explicitly captured by the original reward.

The REINVENT algorithm \citep{olivecrona2017reinvent, blaschke2020reinvent} proposes a reward shaping formulation that couples the agent policy to the prior policy to regularize agent learning of an arbitrary reward function while maintaining important aspects of the prior policy learned, in this case, valid chemistry that resides in a similar chemical space to the prior unsupervised training dataset. This formulation is shown in \autoref{eq:reinvent_reshape}, where $\pi_{prior}$ is the log-likelihood of generating a specific molecule by the prior policy, $\pi_{agent}$ is the log-likelihood of generating a specific molecule by the RL agent, and $\sigma$ is a hyperparameter.

\begin{equation}\label{eq:reinvent_reshape}
\begin{split}
R(\tau)_{reshaped} = \frac{(\pi_{prior} - \pi_{agent} + \sigma \cdot R(\tau))^2}{\pi_{agent}} 
\end{split}
\end{equation}



Here we propose to simplify the REINVENT formulation of REINFORCE while maintaining regularization to the prior policy $\pi_{prior}$. First, we recognize that REINVENT seeks to minimize the difference between the prior and agent policies even if the prior likelihood is low, we alter this behavior such that high prior likelihood is favorable, and low prior likelihood is unfavorable by simply adding the prior policy. Secondly, we decouple the hyperparameter $\sigma$ and assign it to control the effect size of regularization to the prior policy and then we add an exponential coefficient $\alpha$ to investigate effect of increasing the gradient of the reward landscape $R(\tau)$. This is shown in \autoref{eq:reinforce_reg}, note the clip term to ensure the reshaped reward is never negative. 


\begin{equation}\label{eq:reinforce_reg}
\begin{split}
R(\tau)_{reshaped} = clip(R(\tau) + \sigma \cdot \pi_{prior})^{\alpha}
\end{split}
\end{equation}

\textbf{Policy regularization via Kullback-Leibler divergence:}
Alternatively, an additional Kullback-Leibler (KL) divergence loss term can be used to regularize the agent policy (see \autoref{eq:reinforce_kl}). This is a measure of the difference in the distribution of action probabilities given a state between two policies.

\begin{equation}\label{eq:reinforce_kl}
\begin{split}
&\text{KL}(\pi_{prior}\parallel \pi_{agent}) = \\ 
&\lambda_{KL}\cdot\sum _{t=0}^{T} \sum _{a_{i} \in A} \pi_{prior}(a_{i}|s_{t})\ \log \left({\frac {\ \pi_{prior}(a_{i}|s_{t})\ }{\pi_{agent}(a_{i}|s_{t})}}\right)
\end{split}
\end{equation}

\textbf{Exploration via reward shaping:}
The use of diversity filters to penalize molecules already generated can also be considered a form of reward shaping. This could be penalizing repeated episodes or in our context molecules, or penalization by repeated sampling in areas of chemical space recorded by clustering previously generated molecules into different bins via similarity measures \citep{blaschke2020memory, svensson2024diversity}. This is due to the dynamic modification of the reward depending on the output of the diversity filter that changes during optimization. This type of reward shaping can be used to penalize excessive exploitation and hence, encourage exploration into different solution spaces.

\textbf{Exploration via additional loss terms:}
One common method to prevent over-exploitation and to encourage exploration is an entropy (ENT) penalty term as shown in \autoref{eq:reinforce_entropy}. Less common is a high Agent Log-Likelihood (ALL) penalty term to penalize very likely sequences. This is implemented in the original REINVENT algorithm and shown in \autoref{eq:reinforce_all}.

\begin{equation}\label{eq:reinforce_entropy}
\begin{split}
\text{ENT} = -\lambda_{ENT}\cdot\sum _{t=0}^{T} \sum_{a_{i} \in A}  \pi_{\theta}(a_{i}|s_{t}) \log \pi_{\theta}(a_{i}|s_{t})
\end{split}
\end{equation}

\begin{equation}\label{eq:reinforce_all}
\begin{split}
\text{ALL} = -\lambda_{ALL}\cdot\sum_{t=0}^{T}\log \pi_{\theta}(a_t|s_t)^{-1}
\end{split}
\end{equation}

\textbf{Exploration via random network distillation (RND):} RND \citep{burda2018exploration} utilises the error between a target $f$ and estimator neural network $\hat{f}$ to provide a low-overhead bonus intrinsic reward that encourages exploration. The estimator network is trained to estimate the maximum squared error between network outputs given an input $||\hat{f}(x;\theta)-f(x)||^{2}$. This leverages the epistemic uncertainty of neural networks depending on the quantity of training data i.e., few training examples leads to a large error and therefore a large exploration bonus and \textit{vice versa} (\autoref{eq:rnd}). This approach is low overhead because it does not require a memory lookup of previously generated examples as required by diversity filters. This has already proved effective at increasing diversity in CLMs trained with RL \citep{svensson2024diversity}. 

\begin{equation}\label{eq:rnd}
\begin{split}
R(\tau) = R(\tau) + \lambda_{RND}\cdot||\hat{f}(x;\theta)-f(x)||^{2}
\end{split}
\end{equation}

\subsection{Model}
In this work, we utilize a recurrent neural network with gated-recurrent units as the policy model, pre-trained on a subset of molecules extracted from ChEMBL28 \citep{gaulton2012chembl} as described previously \citep{bou2024acegen}. All algorithms were implemented or re-implemented in ACEGEN for consistency. 

\subsection{MolOpt benchmark and performance metrics}
The MolOpt benchmark \citep{guo2022improving} was used as implemented in MolScore \citep{thomas2023molscore} with a budget of 10,000 molecules and optimization was repeated with 5 replicates. This benchmark describes 23 distinct tasks , each associated with different targets. To measure performance, we focus on eight metrics summed over all 23 objectives as described in our previous work \citep{bou2024acegen} (a perfect score for all metrics is 23.). The first four metrics are 'chemistry-naive' proxies for validity, effectiveness (to what extent can the objective score be optimized), efficiency and exploration. The second four 'chemistry-aware' equivalents for the same. 

At the first 'chemistry-naive' level we treat the objective as a perfect description ive where the reward provided covers all aspects of chemical desirability.
\begin{description}[topsep=0pt, parsep=0pt, itemsep=0pt]
    \item[Valid] Validity measured by the proportion of chemically correct molecules.
    \item[Top-10 Avg] Effectiveness measured by the average reward of the best 10 molecules.
    \item[Top-10 AUC] Efficiency measured by the area under the curve of the best 10 molecules cumulatively throughout optimization.
    \item[Unique] Exploration measured by the number of unique molecules generated throughout optimization.
\end{description}

At the second 'chemistry-aware' level we assume an imperfectly described objective that reflects a more realistic perspective for drug discovery. \chadded{Where B\&T-CF refers to 'basic' chemistry filters and includes a set of substructure alerts, LogP range, molecular weight range, number of rotatable bonds and where 'target' chemistry filters include LogP and molecular weight range relative to a target dataset (here the pre-training dataset from ChEMBL28), as well as an ECFP-based outlier analysis. For specific details see Bou et al. \citep{bou2024acegen}.}
\begin{description}[topsep=0pt, parsep=0pt, itemsep=0pt]
    \item[B\&T-CF] Validity measured by the proportion of molecules not considered to have highly idiosyncratic atomic environments or property space relative to the pre-training dataset.
    \item[B\&T-CF Top-10 Avg (Div)] Effectiveness measured by the Top-10 Avg after applying B\&T-CF filters and enforcing that the best 10 molecules represent a diverse selection of chemotypes.
    \item[B\&T-CF Top-10 AUC (Div)] Efficiency measured by the Top-10 AUC after applying B\&T-CF filters and enforcing that the best 10 molecules represent a diverse selection of chemotypes. 
    \item[B\&T-CF Diversity (SEDiv@1k)] Exploration measured by the sphere exclusion diversity metric \citep{thomas2021comparison} of generated molecules after applying B\&T-CF filters. This measures the chemical space coverage of the sample size, here 1,000. A value of 1 can be interpreted as every molecule covering a 'different part' of chemical space, while close to 0 means that all molecules cover the same part of chemical space. 
\end{description}

\section{Results and discussion}\label{results}

\subsection{Exploitation-regularization trade-off}
Maintaining the learnings of a prior policy during RL can help to regulate chemistry to be similar to the prior training dataset which results in the generation of chemistry in desirable chemical spaces. While the reward shaping used by REINVENT helps in practice to regularize learning to the prior policy, the specific formulation we found to be non-intuitive. It is challenging to understand the resulting reward landscape and how to obtain fine-grained control over the trade-off between optimization and regularization. In addition, the numerator is arbitrarily scaled to the power of 2 leading to a steeper gradient in the reward landscape. To better understand the reward landscape from a REINFORCE perspective, we visualized the reward landscape in Supporting Information A. The reward landscape (Figure S1) indicates several curious behaviors. Firstly, in some situations for lower $\sigma$ values, there exists a region of high reshaped reward for very low prior likelihoods and low reward which is actually undesirable. We note that although this idiosyncratic region exists, it may never be encountered in practice at such prior low likelihood values. Secondly, the $\sigma$ value not only affects the shape of the landscape but also the scale and gradient of the reshaped reward leading to one hyperparameter controlling two effects during reshaping. Therefore, we sought an alternative formulation for reward shaping to regularize the REINFORCE algorithm to the prior policy. The outcome of our simplified reshaped reward landscape is visualized in Figure S2, where it can be seen that the shape of the reshaped reward landscape changes with respect to the prior likelihood depending on $\sigma$, the gradient changes depending on $\alpha$ acting independently, while the scale of reshaped reward remains consistent between 0 and 1.

\begin{figure*}[ht]
\begin{center}
\centerline{\includegraphics[width=\textwidth]{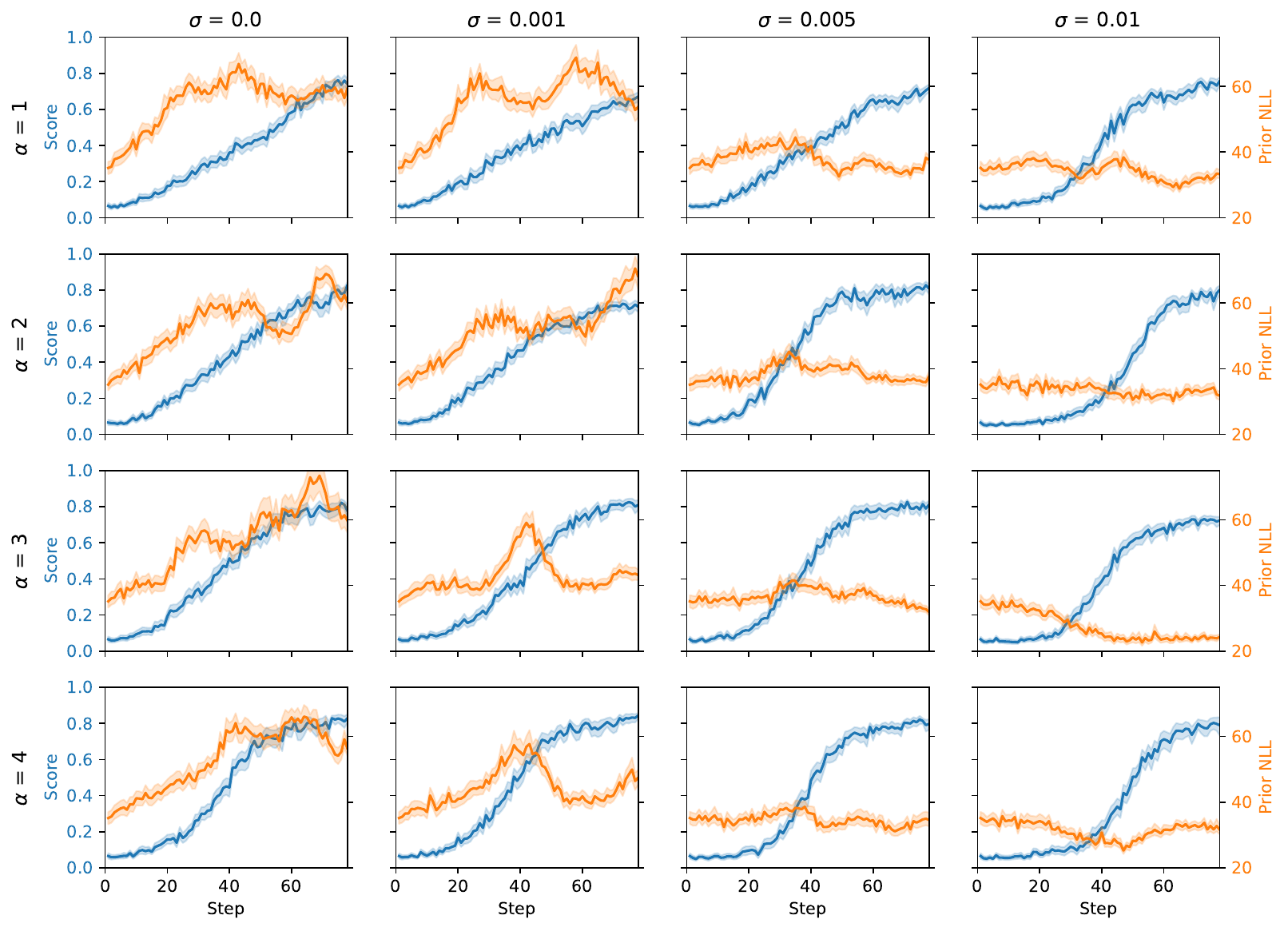}}
\caption{Performance of REINFORCE with the proposed reward shaping at different values of $\alpha$ and $\sigma$ on the JNK3 MolOpt benchmark task. On the left y-axis (blue) is the JNK3 score during the training step, while on the right y-axis (orange) is the negative log-likelihood (NLL) (lower is more likely, and hence better) of molecules according to the prior policy. Variables are measured during RL training steps until 10,000 molecules have been evaluated. Note that higher $\alpha$ increases learning efficiency and higher $\sigma$ increases the prior likelihood, giving control over exploitation-regularization trade-off.}
\label{fig:alpha_sigma}
\end{center}
\end{figure*}

We tested this new reward-shaping mechanism to balance the trade-off between optimization and regularization to a prior policy when conducting RL with REINFORCE. To investigate the behavior of our proposed reward shaping, we titrated the two hyperparameters $\alpha$ and $\sigma$ and conducted RL on an example objective from the MolOpt benchmark, maximization of JNK3 estimated probability of bioactivity. This task is prone to be ``hacking'' where the generative model can drift into undesirable property spaces and generate impractical chemistry \citep{thomas2022re}. The results are shown in \autoref{fig:alpha_sigma}. First, $\sigma$ successfully regularizes RL with increasing values resulting in lower prior negative log-likelihoods (NLL) during training and thus molecules are more likely sampled from the prior policy. Moreover, this remains true at all values of $\alpha$. Meanwhile, increasing values of $\alpha$ successfully results in more efficient optimization, with score optimization shifting from linear to sigmoidal in nature over the budget of 10,000 molecules. Lastly, it can be seen that higher values of $\sigma$ appear to slightly shift the learning curve of the JNK3 score to the right i.e., learning takes slightly longer. The additional constraint to the prior policy effectively adds a potentially inhibitory learning objective. Overall, the reward shaping demonstrates independent and successful control over both return gradient and prior policy regularization not possible with popular alternatives like REINVENT.

\subsection{Extending REINFORCE}

We also evaluated the performance of each REINFORCE extension independently with respect to plain REINFORCE on the MolOpt benchmark to better understand the benefits of each.
 
\begin{figure}[ht]
\begin{center}
\centerline{\includegraphics[width=\columnwidth]{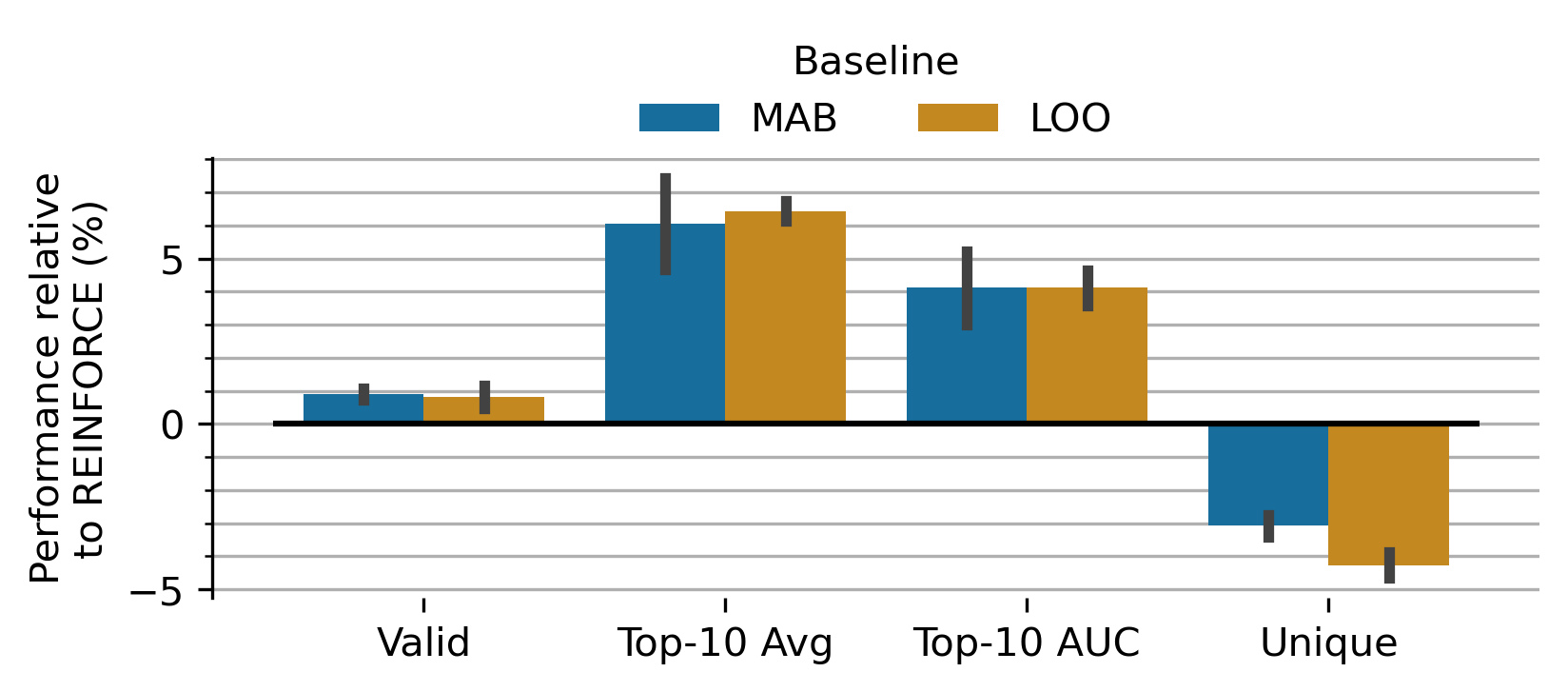}}
\caption{Effect of moving average baseline (MAB) and a leave-one-out baseline (LOO) compared to REINFORCE without baselines. Both baselines increase validity and effectiveness at a small cost to exploration.}
\label{fig:baselines}
\end{center}
\end{figure}

\textbf{REINFORCE baselines:} To reduce the variation in the estimated policy gradients, we investigated the subtraction of a baseline using two methods 1) subtraction of a moving average baseline and 2) subtraction of a leave-one-out baseline. \autoref{fig:baselines} shows that the subtraction of either of these baselines improves the effectiveness by $\sim 6\%$ and efficiency by $\sim 4\%$. This comes at a slight compromise with a drop in the number of unique molecules of $\sim 3\%$, indicating less exploration which is likely a factor of enhanced optimization. Overall the moving average baseline appears to show the best overall profile for performance improvement over REINFORCE.

\begin{figure}[ht]
\begin{center}
\centerline{\includegraphics[width=\columnwidth]{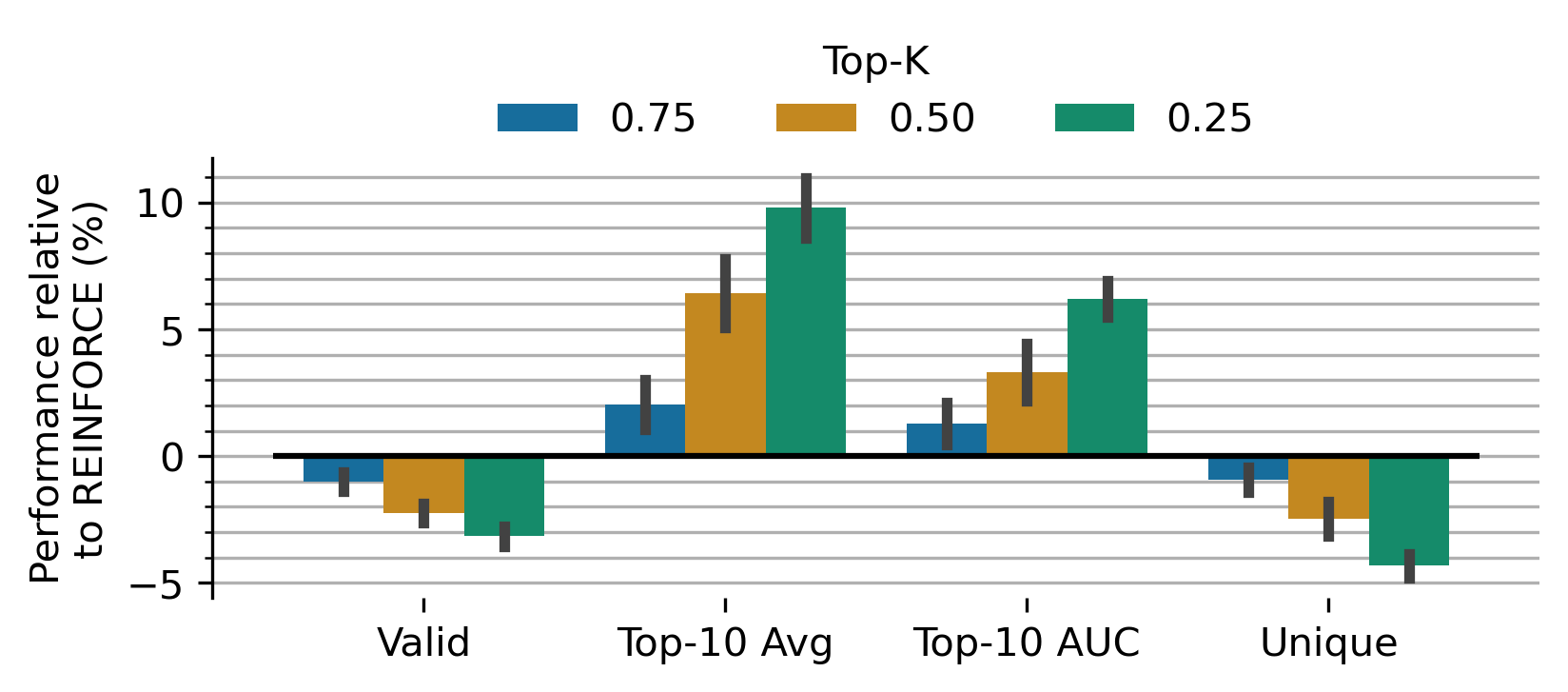}}
\caption{Effect of different top-k on-policy subset values (hill-climbing) compared to REINFORCE (which corresponds to top-k=1). Decreasing subset sizes considerably increases effectiveness and efficiency at smaller relative cost to validity and uniqueness.}
\label{fig:hill_climb}
\end{center}
\end{figure}

\textbf{Hill-climb:} Hill-climb is a strategy that subsets the highest scoring molecules on-policy; we evaluate different top-k values and their effect on learning efficiency shown in \autoref{fig:hill_climb}. We observe that as the top-k ratio decreases, the effectiveness and efficiency increase correspondingly. This strategy can considerably boost effectiveness as much as $\sim 10\%$ and efficiency as much as $\sim 6\%$ with only small drops in validity and uniqueness of $\sim 3\%$ and $\sim 4\%$ respectively.

\begin{figure}[ht]
\begin{center}
\centerline{\includegraphics[width=\columnwidth]{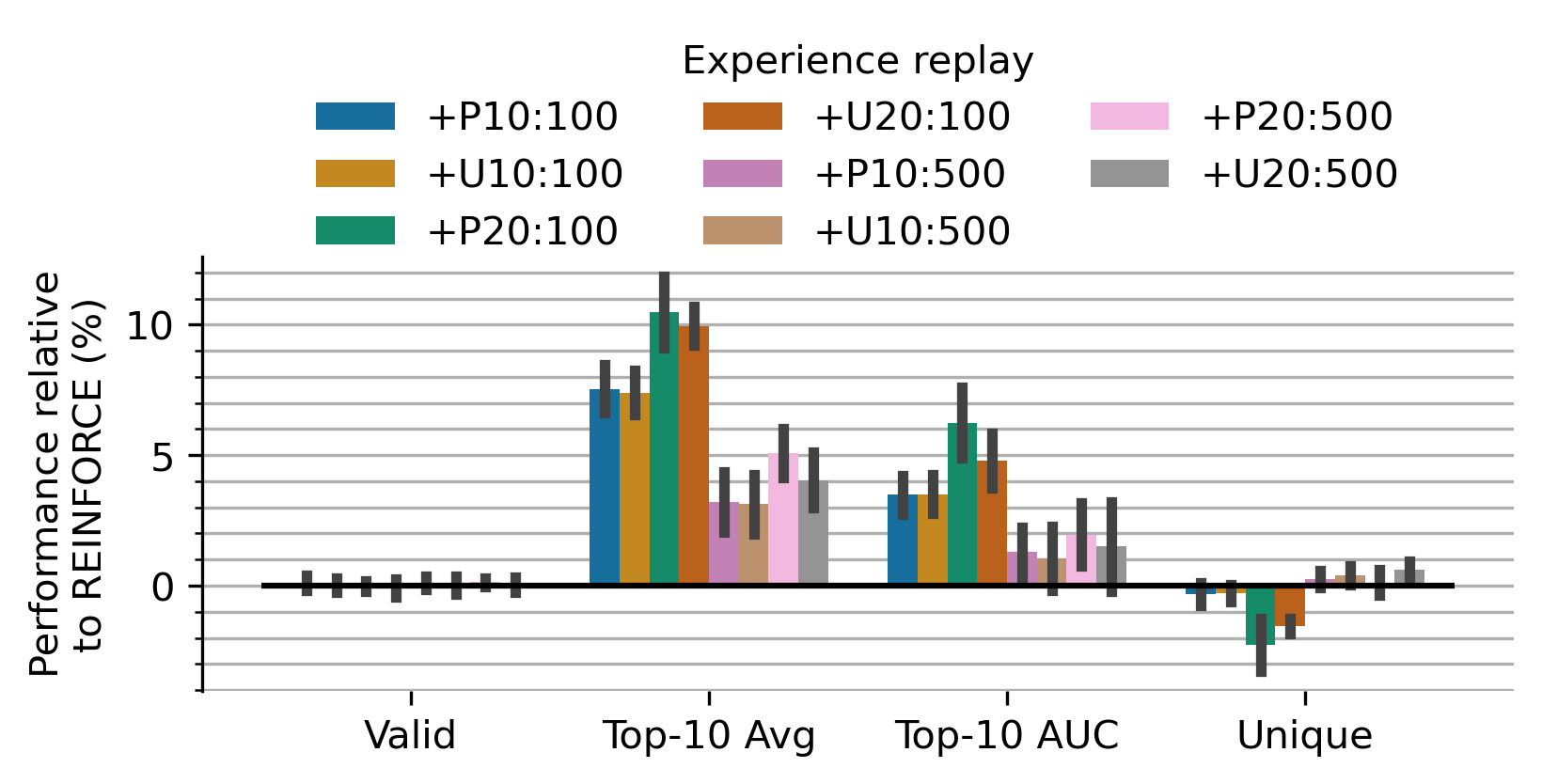}}
\caption{Effect of different experience replay sampling strategies compared to REINFORCE with no experience replay. The initial letter indicates the sampling type: prioritized proportional to the molecule's reward (P) or uniform (U). The first number after the letter represents the replay batch size, and the final number indicates the experience replay buffer size. Most experience replay strategies increase effectiveness and efficiency at no cost to validity or uniqueness, with smaller replay buffers having a greater effect.}
\label{fig:experience_replay}
\end{center}
\end{figure}

\textbf{Experience replay:} We explored several experience replay configurations to augment the learning process with off-policy data. Molecules generated by the agent with the highest rewards are stored in a replay buffer and two buffer sampling methods are evaluated to augment on-policy data: sampling buffer molecules uniformly or proportionally to their reward. Additionally, we consider two different sizes for the replay buffer (100 and 500 molecules) and two different batch sizes for sampled buffer data (10 and 20). The results, displayed in \autoref{fig:experience_replay}, indicate that using replayed off-policy data consistently enhances effectiveness (up to $\sim10\%$) and sample efficiency (up to $\sim6\%$), while having no effect on the number of valid molecules generated and limited effect on exploration as measured by the proportion of unique molecules. Prioritized sampling leads to larger performance increases, however, uniform sampling marginally lessens the effect of decreased uniqueness. We find that, out of the configurations tested, a buffer size of 100, batch size of 20 and proportional sampling maximizes both performance and efficiency.

\begin{figure}[ht]
\begin{center}
\centerline{\includegraphics[width=\columnwidth]{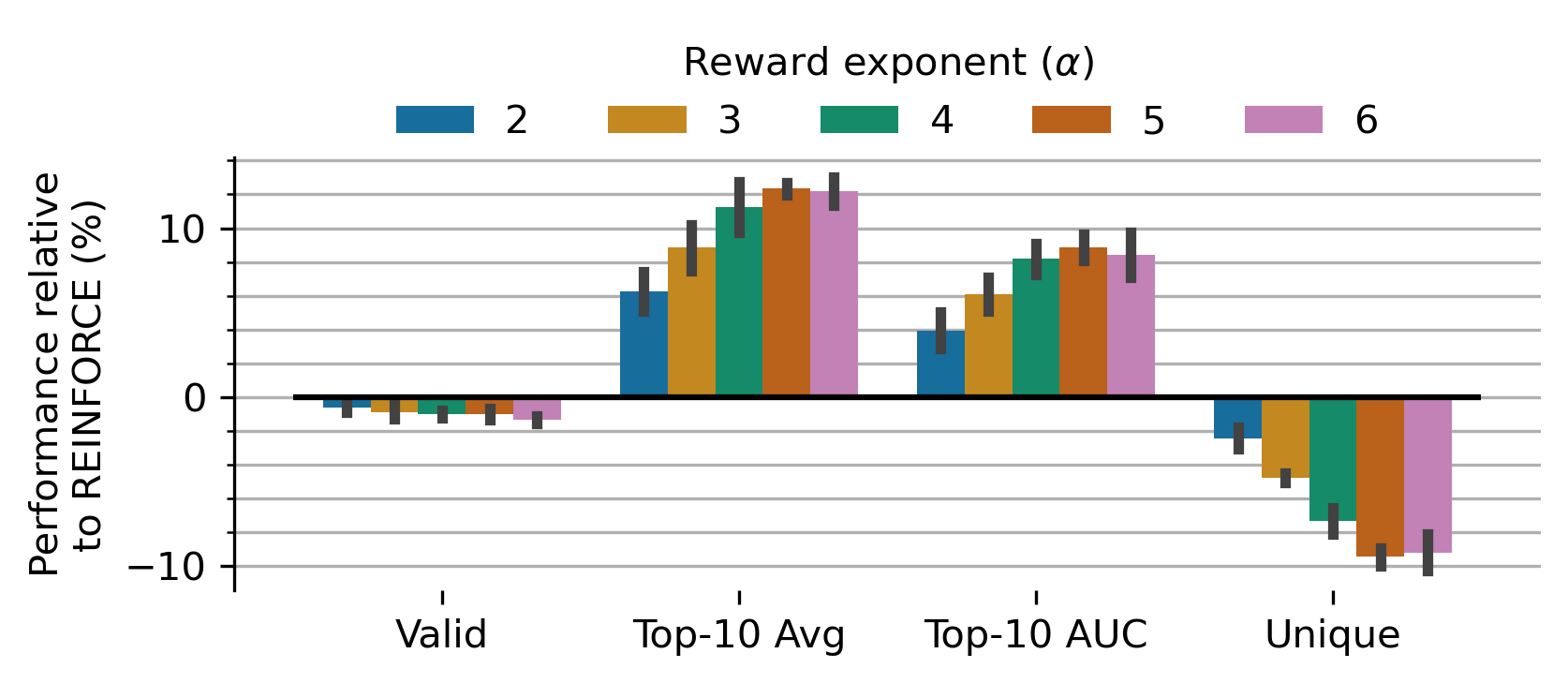}}
\caption{Effect of different reward exponent values compared to REINFORCE (which corresponds to a value of 1). Increasing exponent values that steepen the return gradient lead to increased effectiveness and efficiency at a cost to exploration.}
\label{fig:reward_exp}
\end{center}
\end{figure}

\textbf{Reward exponent:} The new reward shaping mechanism proposed here introduces a tune-able reward exponent $\alpha$. Note that REINVENT uses a fixed exponential function with a power of 2. To better understand how this exponentiation affects the optimization process, we experimented with several exponent values. \autoref{fig:reward_exp} shows that increasing the exponent value consistently improves the effectiveness and efficiency up to $\sim12\%$ and $\sim9\%$ before saturating at an exponent of 5, beyond which no further benefit is observed. Although this reward shaping has a minimal impact on the proportion of valid molecules, it does correspondingly reduce observed exploration, reducing the proportion of unique molecules generated up to $\sim9\%$.

\begin{figure}[ht]
\begin{center}
\centerline{\includegraphics[width=\columnwidth]{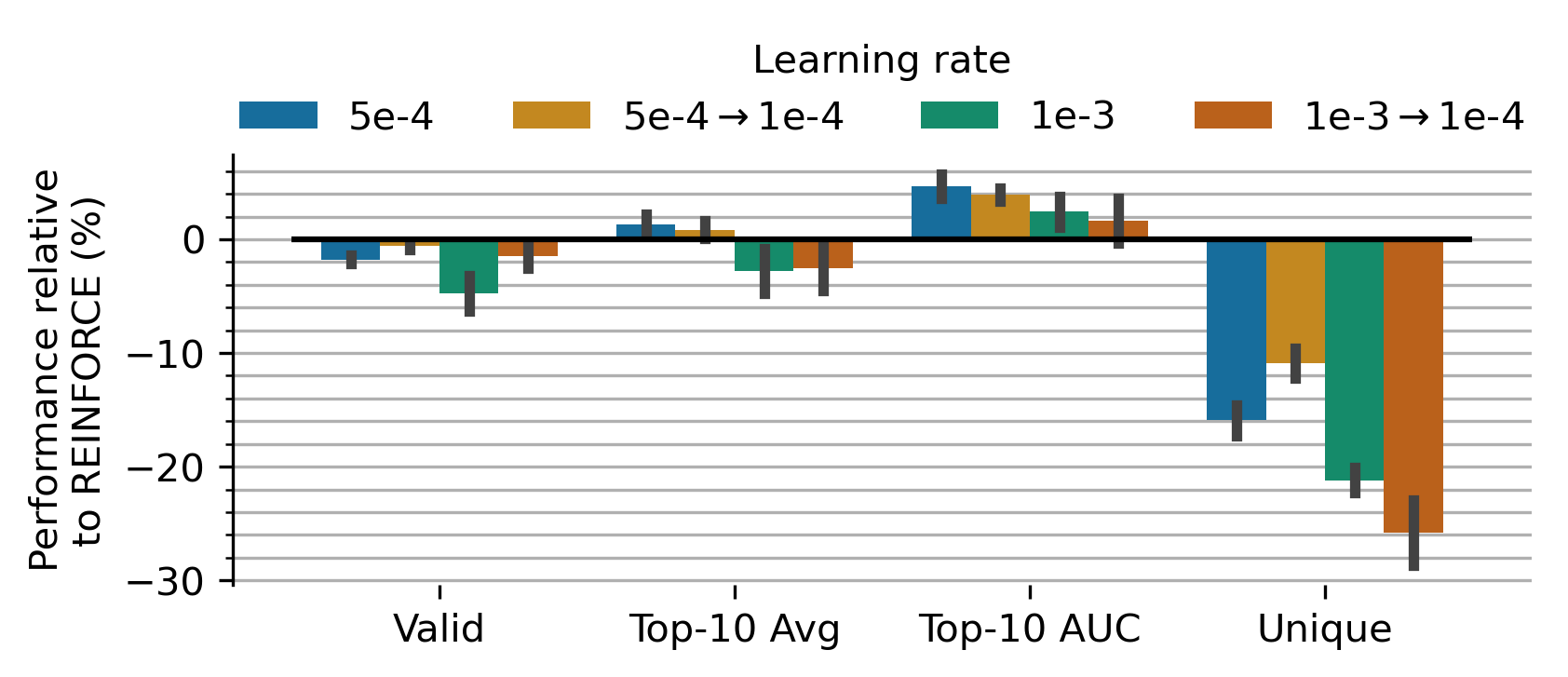}}
\caption{Effect of different learning rates compared to REINFORCE (where the default learning rate is 1e-4). The arrow represents the cosine annealing from 5e-4 to 1e-1 and from 1e-3 to 1e-4. Higher learning rates show marginal benefit in learning efficiency at a high cost to exploration.}
\label{fig:learning_rate}
\end{center}
\end{figure}

\textbf{Learning rate:} The learning rate of the optimizer could also afford more efficient learning analogous to the update step size in the REINFORCE policy. \autoref{fig:learning_rate} shows the effect of increasing the learning rates from the default 1e-4, as well as annealing the learning rate over the first half of optimization back to the default 1e-4. We see that increasing the learning rate to 5e-4 increases efficiency but only up to $\sim5\%$, but further increases lead to a decrease in efficiency. Expectedly, all increases result in significantly less exploration as much as $\sim26\%$. Annealing the learning rate from 5e-4 to 1e-4 does ameliorate the collapse in exploration to $\sim10\%$ for an efficiency gain of $\sim4\%$. Overall, this limited set of changes in learning rate does not seem to provide noteworthy performance gains.
 
\begin{figure*}[ht]
\begin{center}
\centerline{\includegraphics[width=\textwidth]{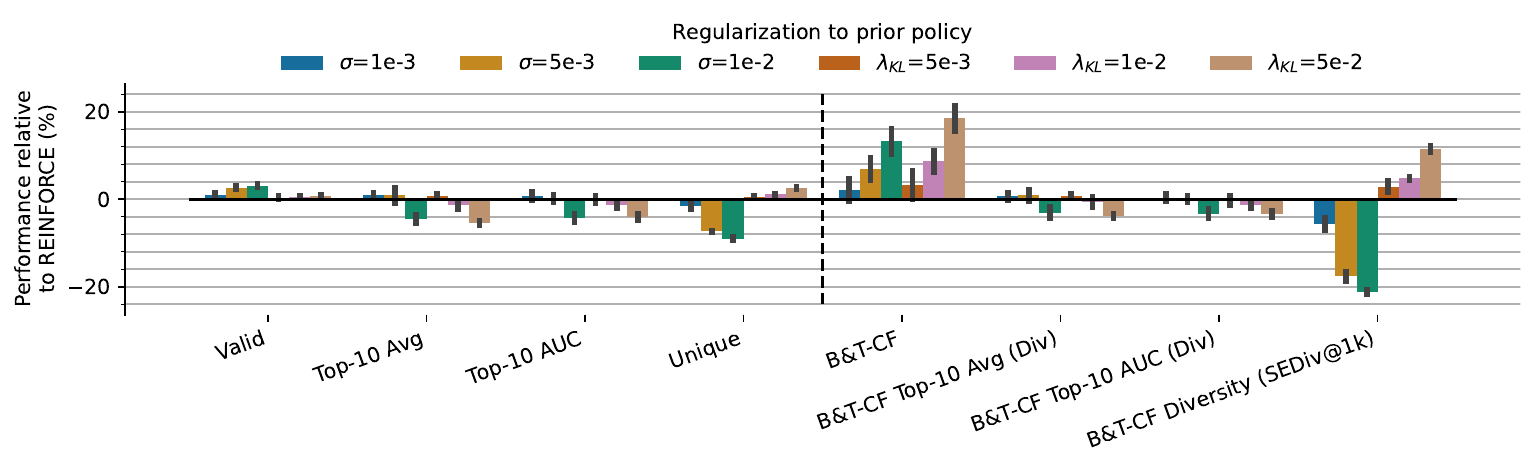}}
\caption{Effect of different policy regularization strategies and parameters, including different $\sigma$ for the proposed reward shaping and different $\lambda_{KL}$ coefficients for regularization by KL divergence. Increasing coefficients of both strategies improve validity, more profoundly observed by the 'chemistry-aware' metrics on the right. However, the proposed reward shaping correspondingly decreases exploration, whereas KL divergence correspondingly increases exploration.}
\label{fig:regularization}
\end{center}
\end{figure*}

\textbf{Prior regularization:} Regularizing the agent policy to the prior policy can not only avoid catastrophic forgetting but also ensure important aspects of the prior policy are maintained such as the property and chemistry distribution of the training set. In addition to regularization via our proposed reward shaping mechanism, we tested the KL divergence between the agent and prior log-likelihoods. We explore various values for the regularization coefficient $\sigma$ in our proposed reward reshaping mechanism, as well as different KL coefficients \chadded{($\lambda_{KL}$)}, illustrated in \autoref{fig:regularization}. Note that we did not test different $\sigma$ for the REINVENT as this does not independently control regularization, but also the scale and gradient of the return landscape. As these parameters more directly relate to the quality of chemistry output, we additionally show the more 'chemistry-aware' metrics that also measure the quality of chemistry. Most notably, we observed that the addition of our reward shaping positively impacts the validity of proposed molecules up to $\sim12\%$ but does decrease the diversity of the output by up to $\sim20\%$: likely due to the constraint of prior likelihood put on the agent, limiting freedom of exploration. Meanwhile, the KL divergence loss term also increases chemical validity up $\sim~18\%$ and increases exploration up to $\sim~12\%$. This effect is likely due to the KL term penalizing diverging \textit{distributions} of actions at each timestep relative to the prior which will have higher entropy than a focused agent. Whereas, any strategy that only accounts for the overall likelihood does not model such distributional changes. However, both strategies lead to a small drop in effectiveness and efficiency. These results show that in this case the KL divergence is more beneficial to regularize prior policy than reward shaping with prior log-likelihood values.

\begin{figure*}[ht] 
\begin{center}
\centerline{\includegraphics[width=\textwidth]{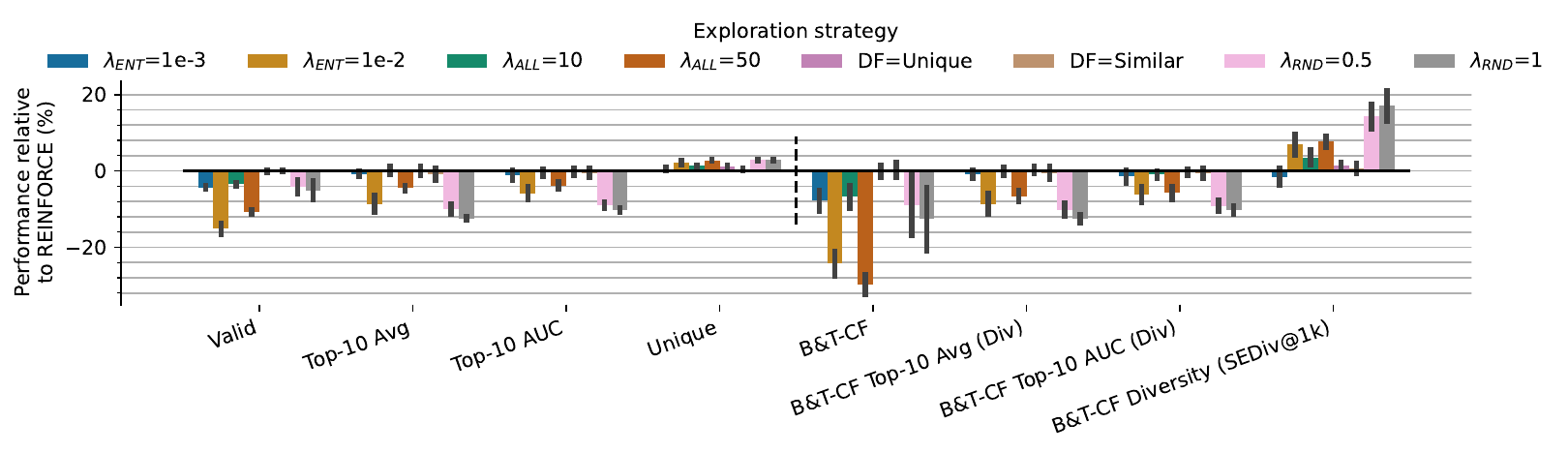}}
\caption{Effect of different exploration strategies and parameters, including an entropy loss term (ENT), a high agent likelihood penalty (ALL), two diversity filters (DF), and random network distillation (RND). Note that all strategies increase exploration measured by B\&T-CF Diversity to varying degrees, but typically come at the cost of decreased validity, effectiveness and efficiency.}
\label{fig:exploration}
\end{center}
\end{figure*}

\textbf{Exploration techniques:} Most of the techniques analyzed so far improve optimization ability but often at the cost of uniqueness, our proxy for exploration. Here, we test strategies to increase agent exploration on the more rigorous 'chemistry-aware' metrics including exploration measured by chemical diversity. We considered four strategy options: Maximizing the entropy (ENT), penalizing highly likely agent sequences (ALL), two diversity filters (DF) that either penalize the repetition of molecules, or that penalize molecules similar to those generated previously, and finally a random network distillation (RND) bonus exploration reward. \chadded{We test ENT, ALL, and RND with two different coefficients ($\lambda$) in the loss function} (see Equations \ref{eq:reinforce_entropy}, \ref{eq:reinforce_all}, \ref{eq:rnd}). \autoref{fig:exploration} shows that all strategies except the diversity filters increase exploration via B\&T-CF Diversity up to $\sim18\%$ with RND. However, all strategies except for diversity filters do this at the cost of validity (up to $\sim28\%$), effectiveness (up to $\sim12\%$), and efficiency (up to $\sim12\%$). Higher values of $\lambda_{ENT}$ and $\lambda_{ALL}$ were tested but led to considerably worse performance (see Supporting Information D). We find that RND provides the best trade-off in diversity increased vs validity dropped, which may be mitigated by combination with policy regularization strategies. Furthermore, the diversity filters have little impact on performance in this benchmark, due to the short budget of 10,000 molecules not filling memory buffers enough to penalize rewards and affect exploration. We refer the reader to other studies of diversity filters \citep{blaschke2020memory, thomas2022augmented, svensson2024diversity} for a more thorough investigation of this strategy. We also propose that REINFORCE is already reasonably explorative and these strategies are more beneficial when combined with the other extensions to counter the decreases in exploration observed.

\subsection{Optimizing REINFORCE}

The extensions to REINFORCE individually tested here are likely not orthogonal and may be anti-correlated. To search for an optimal combination of hyperparameters, we conducted a hyperparameter search based on the most promising extensions observed (see Supporting Information C). We tested a random subset of 1,000 configurations out of a combinatorially possible 7,464,960 and approximated performance with two tasks from the MolOpt benchmark. The two tasks selected were "Osimertinib MPO" and "Median molecules 2" that most correlated with overall benchmark performance. Where the best-performing configuration was the one with the highest Top-10 AUC as proposed in the original publication. 

The configuration with the best performance observed during hyperparameter optimization we call ACEGEN$_{MolOpt}$, with the corresponding results shown in \autoref{tab:combined_performance}. This configuration shows state-of-the-art performance in effectiveness (Top-10 Avg) and efficiency (Top-10 AUC). \chadded{Although a Bonferroni corrected Wilcoxon signed-rank test on the aggregated benchmark results only showed significant improvement in validity (p=0.002 compared to REINVENT-MolOpt), an analysis at the task level with a TukeyHSD considering all replicates revealed that ACEGEN$_{MolOpt}$ significantly outperformed REINVENT-MolOpt at several tasks in Top-10 Avg and Top-10 AUC (Figure S5)}. Similar to REINVENT$_{MolOpt}$, ACEGEN$_{MolOpt}$ particularly suffers from low exploration and low B\&T-CF validity. In the event that the scoring function(s) perfectly describes chemical desirability, this is not a problem. Moreover, this configuration was identified by searching only $\sim0.01\%$ of the combinatorial space. Therefore, it is likely a better-performing configuration exists. We make the hyperparameter search scripts available if further search of this space is of interest to researchers.

On the other hand, more often than not scoring functions do not perfectly describe and account for chemical desirability. In this case, we hand-selected a configuration we call ACEGEN$_{Practical}$ to better maintain chemical validity and exploration, for which the hyperparameters are shown in Supporting Information B. This configuration achieves a balance across all metrics. Maintaining similar B\&T-CF validity to AHC but improving markedly in effectiveness and efficiency.

\begin{table*}[h!]
\centering
\resizebox{\linewidth}{!}{
\begin{tabular}{|l|cccc|cc|}
\hline
 & \multicolumn{4}{c|}{References} & \multicolumn{2}{c|}{Ours} \\
Metric & REINFORCE & REINVENT & REINVENT-MolOpt & AHC & ACEGEN\(_{Practical}\) & ACEGEN\(_{MolOpt}\) \\ 
\hline
Valid & 21.77 ± 0.03 & 21.79 ± 0.02 & 21.74 ± 0.05 & 21.45 ± 0.02 & 22.03 ± 0.03 & \textbf{22.17 ± 0.04} \\
Top-10 Avg & 15.85 ± 0.10 & 15.67 ± 0.11 & 17.43 ± 0.17 & 16.09 ± 0.10 & 16.82 ± 0.21 & \textbf{17.63 ± 0.13} \\
Top-10 AUC & 13.67 ± 0.07 & 13.55 ± 0.08 & 15.65 ± 0.14 & 13.91 ± 0.08 & 14.27 ± 0.12 & \textbf{15.94 ± 0.09} \\
Unique & 22.28 ± 0.10 & 22.67 ± 0.04 & 13.68 ± 0.31 & \textbf{22.68 ± 0.07} & 18.21 ± 0.30 & 10.39 ± 0.34 \\
\hline
B\&T-CF & 14.34 ± 0.18 & \textbf{14.70 ± 0.12} & 7.00 ± 0.27 & 13.77 ± 0.12 & 13.30 ± 0.25 & 6.18 ± 0.27 \\
B\&T-CF Top-10 Avg (Div) & 14.95 ± 0.12 & 14.79 ± 0.13 & 16.06 ± 0.16 & 15.25 ± 0.10 & 15.50 ± 0.20 & \textbf{16.07 ± 0.18} \\
B\&T-CF Top-10 AUC (Div) & 12.91 ± 0.06 & 12.81 ± 0.08 & 14.61 ± 0.13 & 13.11 ± 0.08 & 13.28 ± 0.14 & \textbf{14.72 ± 0.14} \\
B\&T-CF Diversity (SEDiv@1k) & 17.39 ± 0.15 & \textbf{18.23 ± 0.11} & 10.10 ± 0.27 & 17.55 ± 0.10 & 15.69 ± 0.16 & 12.54 ± 0.25 \\
\hline
\end{tabular}}
\caption{Performance comparison between proposed ACEGEN RL algorithms and baseline algorithms on the MolOpt benchmark. ACEGEN$_{Practical}$ provides a hand-picked configuration aimed to balance validity, effectiveness, efficiency and exploration; this outperforms its corresponding baseline AHC. ACEGEN$_{MolOpt}$ provides a hyperparameter optimized configuration for the MolOpt benchmark focusing on effectiveness and efficiency, outperforming its corresponding baseline REINVENT$_{MolOpt}$.}
\label{tab:combined_performance}
\end{table*}

\subsection{Case study: Optimizing putative JNK3 binding affinity via Boltz2}

\begin{figure*}[ht] 
\begin{center}
\centerline{\includegraphics[width=\textwidth]{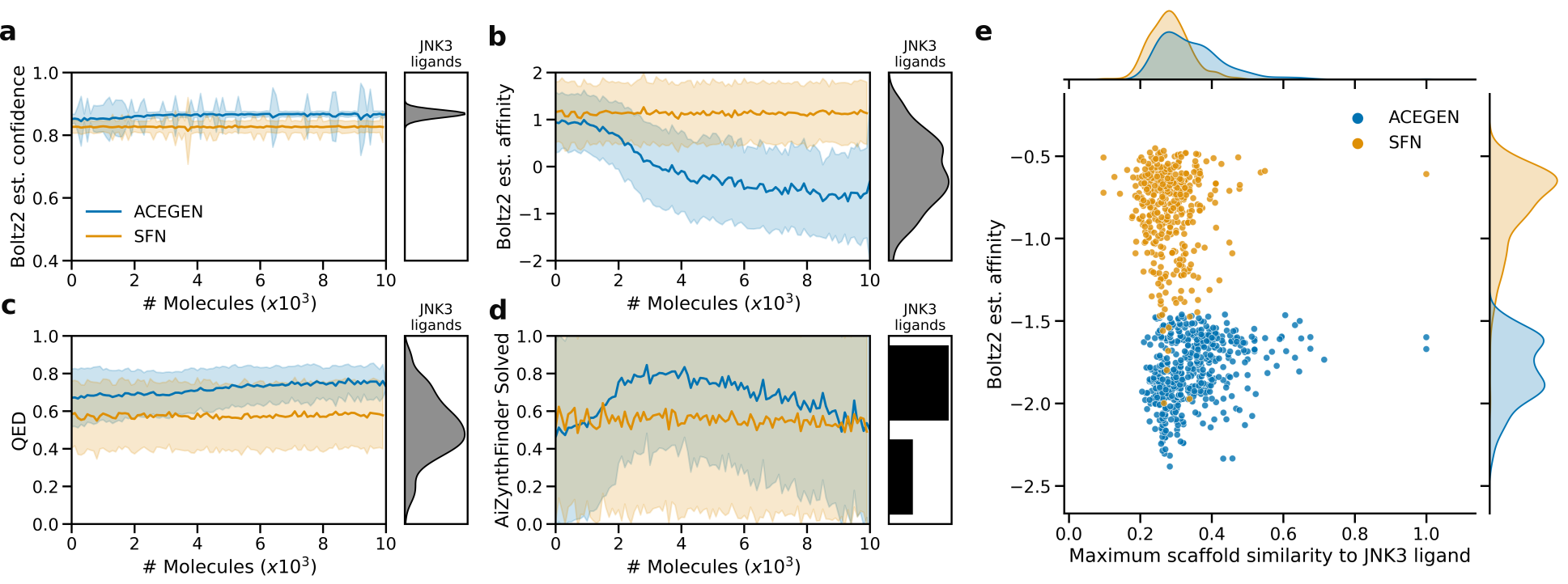}}
\caption{Comparison between ACEGEN and SynFlowNet generated molecules when optimizing the Boltz2 binding affinity of JNK3. These experiments were repeated 5 times with the mean and standard deviation shown. (a) The binding confidence remains high during learning. (b). The affinity improves (decreases) as ACEGEN learns, however, SynFlowNet doesn't improve within the budget. (c) The QED is improved more so by ACEGEN. (d) \chreplaced{Proportion synthesizable as determined by finding a synthetic route with AiZynthFinder}{Synthesizability score (SAScore) remains equally low (lower is better) for both models}. Reference values for known JNK3 ligands are shown for each property \chadded{Either solved=1, or not solved=0}. (e) Distribution of estimated binding affinity and scaffold similarity (Tanimoto similarity of Bemis-Murcko scaffolds by ECFP4 fingerprints) to JNK3 ligands of the top 500 molecules from each model. Neither model generated very similar molecules to known JNK3 ligands.}
\label{fig:JNK3_baseline_opt}
\end{center}
\end{figure*}

To better assess the capability of the ACEGEN$_{MolOpt}$ RL configuration (now on referred to as ACEGEN) we tested its ability to generate molecules guided by frontier binding affinity models \citep{passaro2025boltz}. This was also done within a practical evaluation budget of 10,000 molecules and compared to SynFlowNet \citep{cretu2024synflownet} as the baseline model used by Passaro et al. \cite{passaro2025boltz}. SynFlowNet is a GFlowNet generative model that is constrained to synthesizable chemical space, hence increasing practicality for industrial drug discovery but also restricting chemical space explored. \autoref{fig:JNK3_baseline_opt} shows that ACEGEN is able to optimize Boltz2 binding affinity well within the budget, which SynFlowNet does not manage to achieve. This is not especially surprising considering Passaro et al. evaluated up to 400,000 molecules until policy convergence. As SynFlowNet is restricted to synthesizable chemical space, we expected \chreplaced{more molecules to be solved by AiZynthFinder \citep{genheden2020aizynthfinder}. However, we found the opposite, that proportionally ACEGEN had more molecules with successfully identified synthetic routes throughout optimization. We hypothesize that this is likely due to chemical space biases and alignment between the ChEMBL pre-trained CLM and AiZynthFinder, compared to SynFlowNet which uses Enamine building blocks and reaction sets. However, it is encouraging that ACEGEN-proposed designs are also predicted to be synthesizable.}{the synthetic accessibility of molecules by SAScore \citep{ertl2009estimation} to be significantly lower (better) than ACEGEN generated molecules. However, molecules from both remain equally low through learning.} \chadded{For examples of molecules with predicted solved or unsolved synthetic routes see Figure S6 and S7.} We did not observe that molecules from either model, or those with better estimated binding affinity were more close to known JNK3 ligands. This suggests that the models are probing novel chemical spaces relative to JNK3 ligands. Overall, ACEGEN is much more capable of learning than SynFlowNet for this particular task and budget, and remains in desirable chemical spaces as measured by QED \citep{bickerton2012quantifying} and SAScore. 

\begin{figure*}[ht] 
\begin{center}
\centerline{\includegraphics[width=\textwidth]{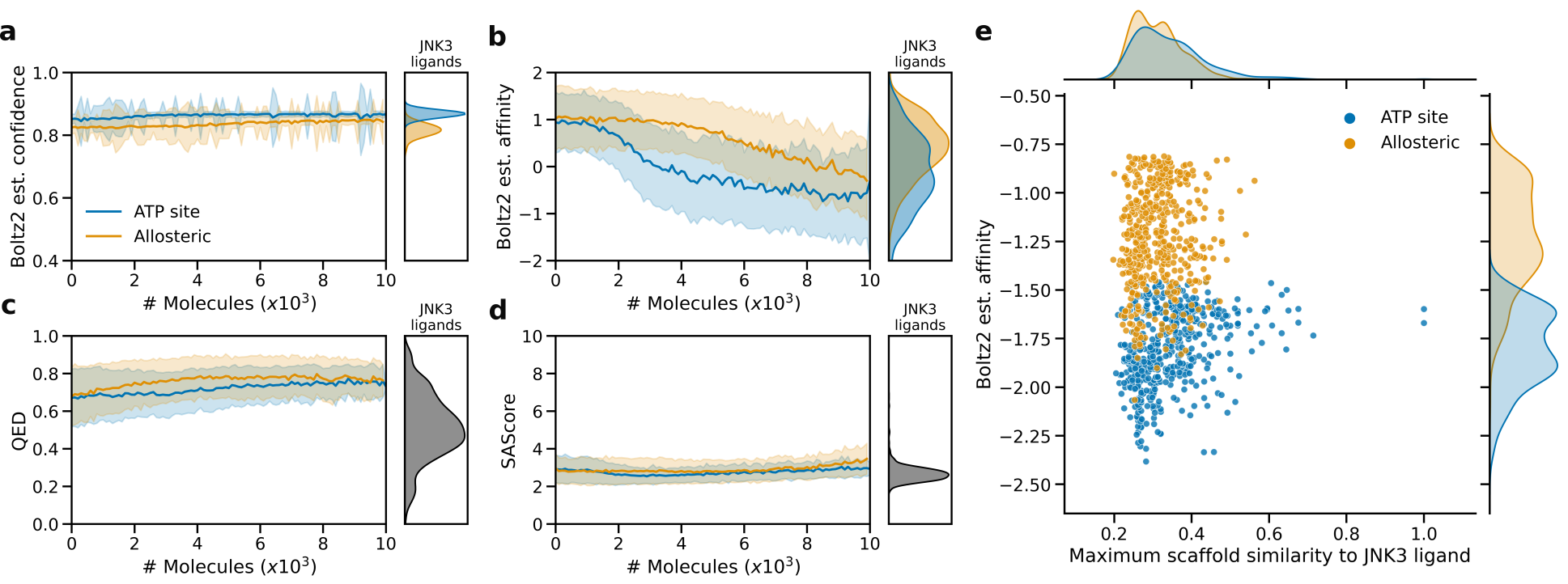}}
\caption{Comparison between optimizing Boltz2 binding affinity in the ATP site or an allosteric site by pre-ocupation of ATP in the ATP site. These experiments were repeated 5 times with the mean and standard deviation shown. (a) The binding confidence remains high during learning. (b). Estimated binding affinity improves at a slower rate for allosteric binding. (c) The QED remains high for both tasks. (d) Synthesizability score (SAScore) remains equally low for both tasks. Reference values for known JNK3 ligands are shown for each property. (e) Distribution of estimated binding affinity and scaffold similarity (Tanimoto similarity of Bemis-Murcko scaffolds by ECFP4 fingerprints) to JNK3 ligands of the top 500 molecules from each task. Neither task generates very similar molecules to known JNK3 ligands.}
\label{fig:JNK3_allo_opt}
\end{center}
\end{figure*}

\begin{figure*}[ht] 
\begin{center}
\centerline{\includegraphics[width=0.8\textwidth]{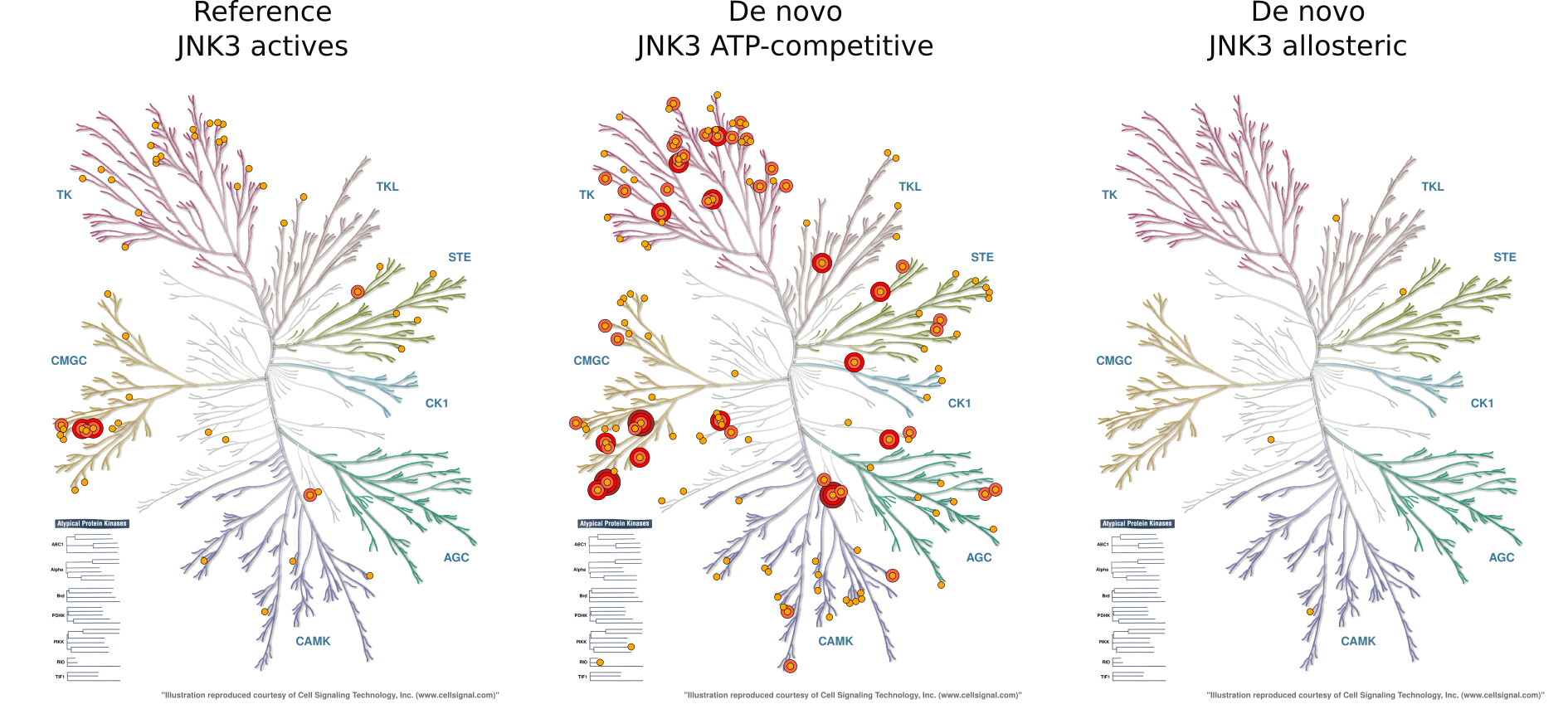}}
\caption{Estimated kinase selectivity profile for JNK3 ligands, ATP-competitive de novo molecules, and Allosteric de novo molecules. Kinases are annotated according to the percent of ligands estimated active at a concentration of 10 $\mu$M (dark red >80\%, red >60\%, salmon >40\%, orange >20\%). Figures were generated using KinMap\citep{eid2017kinmap}.}
\label{fig:JNK3_allo_sel}
\end{center}
\end{figure*}

A pivotal challenge in finding kinase inhibitor drug candidates is selectivity, due to highly conserved ATP-binding sites across protein kinases. Allosteric ligands targeting protein pockets show particular promise in identifying more selective drugs \citep{fang2013strategies}. Therefore, we ran an additional experiment to optimize the Boltz2 binding affinity whilst additionally co-folding ATP into the orthosteric pocket, encouraging the model to fold putative ligands elsewhere in potential allosteric pockets. \autoref{fig:JNK3_allo_opt} shows that learning to optimize the estimated binding affinity is slower but possible, while still maintaining high confidence, and desirable QED and SAScore property value ranges. Likewise, the top 500 molecules (top 100 from each experimental repeat) do not bare close resemblance to known JNK3 ligands - which are mostly orthosteric, ATP-competitive ligands. To estimate kinase selectivity of the resulting molecules, we estimated the binding probability of each molecule to 430 different kinases across the kinome. This was done by utilising the publicly available random forest models trained to classify ligands taken from ChEMBL  and PubChem for each kinase at a concentration of 10 $\mu$M \citep{thomas2023pidginv5}. \autoref{fig:JNK3_allo_sel} confirms that the de novo molecules targeting allosteric sites have much improved estimated selectivity profiles than known JNK3 ligands. On the other hand, the ATP-competitive de novo molecules have much worse estimated selectivity profiles than known JNK3 ligands. Therefore, ATP-competitive de novo molecules appear to contain pan-kinome chemotypes which may be influenced by Boltz2 encouraging general kinase ligand features as opposed to, JNK3 specific features.

\begin{figure*}[ht] 
\begin{center}
\centerline{\includegraphics[width=\textwidth]{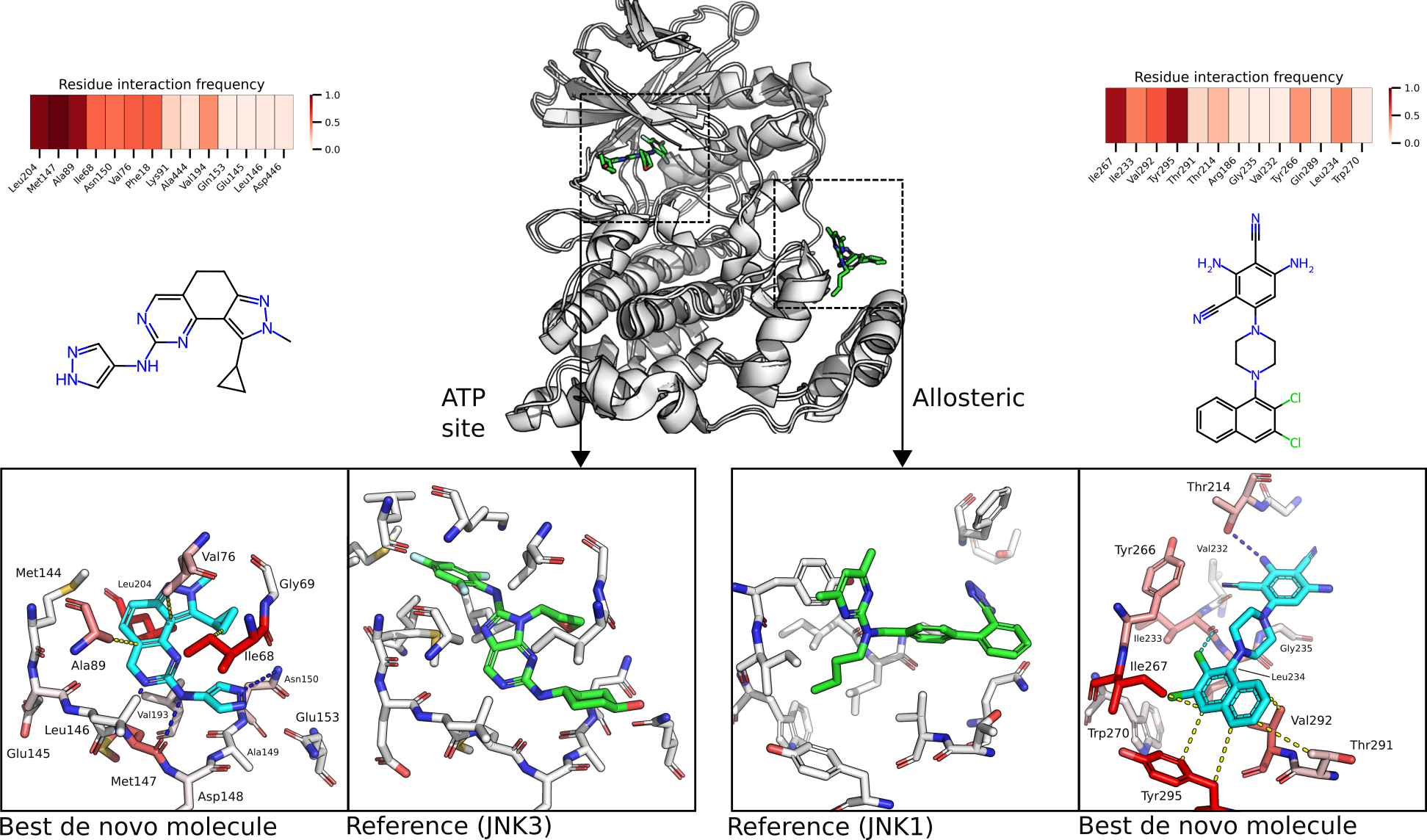}}
\caption{De novo molecule with the best Boltz2 binding affinity for each ATP-competitive and allosteric experiment, alongside reference structures. JNK3 reference is the CC-930 drug candidate (PDB: 3TTI) \citep{krenitsky2012discovery}. There are no reported JNK3 small-molecule allosteric structures, hence, the reference structure is JNK1 isoform bound to an allosteric inhibitor (PDB:3O2M)\citep{comess2011discovery} in the DEF site \citep{jacobs1999multiple}.}
\label{fig:JNK3_allo_3D}
\end{center}
\end{figure*}

Inspecting the co-folded structures showed that the vast majority of putative allosteric ligands were co-folded into the docking site for ERK or FXF (DEF) \cite{jacobs1999multiple}. 
\chadded{This is a substrate-binding site located between the MAPK insert and $\alpha$G helix that recognizes conserved FXF motifs required for binding by some downstream substrates, although better characterized in ERK \cite{jacobs1999multiple} this motif has been implicated in MAPK phosphatases determining JNK1 recognition \cite{liu2016conserved}.}\chdeleted{This allosteric site is found in some MAPK kinases including some JNK kinases.} However, there is no publicly available structure of a small molecule bound to the DEF site of JNK3, there only exists one of the closely related JNK1 kinase \cite{comess2011discovery}. The best de novo compounds for each pocket are shown in \autoref{fig:JNK3_allo_3D}. The best allosteric compound is projected to bind with a di-chloro napthalene moiety in the shallow lipophilic pocket forming hydrophobic interactions with Tyr295, Thr291, Val292, and Ile267 with an additional halogen bond to the backbone carbonyl of Ile233. Conversely, the rest of the compound is polar with a piperazine core and 2,4-dicyano-1,3-diaminobenzene forming a hydrogen bond with Thr214. Although the 2,4-dycyano-1,3-diaminobenze structure is likely not synthetically accessible or stable in this case, many other generated compounds share this amiphiphillic nature with the piperazine core commonly appearing (see Supporting Information E.3).

\begin{figure*}[ht]
\begin{center}
\centerline{\includegraphics[width=\textwidth]{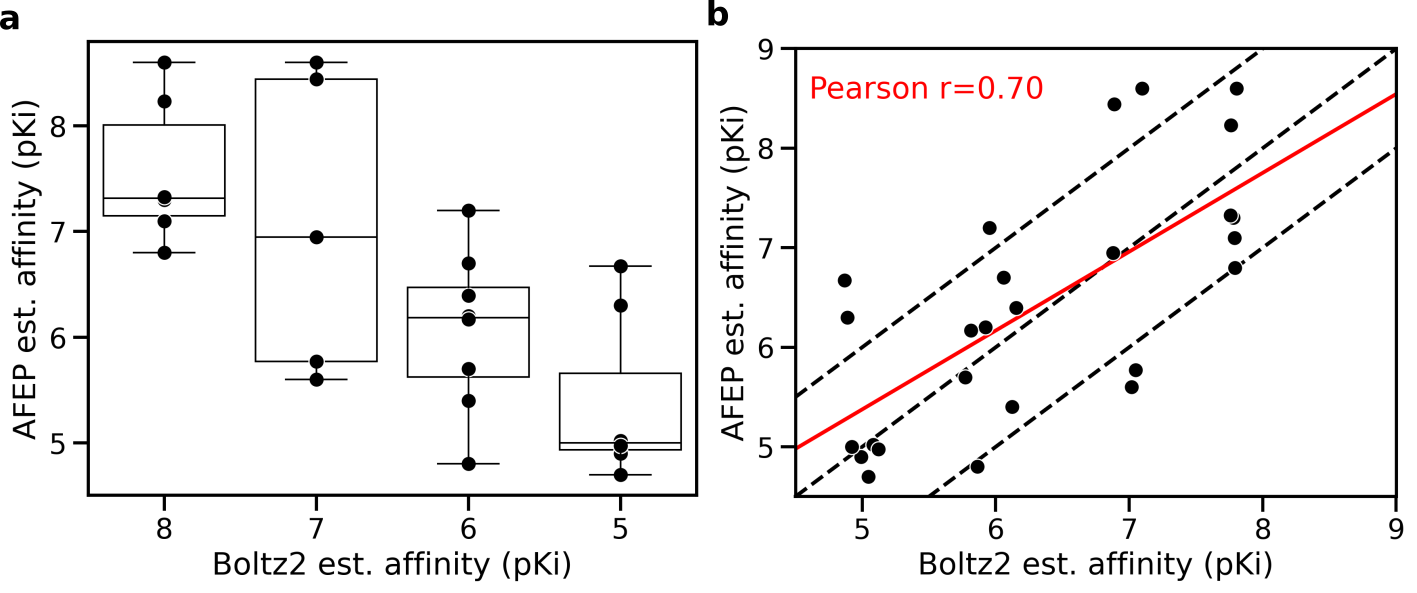}}
\caption{Correlation between Boltz2 estimated pKi and ABFE estimated pKi. (a) Binned affinities show a positive stepwise correlation, where a one-way ANOVA test reports significant variance between the means (p=0.0012). (b) The non-binned correlation between Boltz2 estimated affinity and AFEP with a reported Pearson correlation of 0.7.}
\label{fig:AFEP_validation}
\end{center}
\end{figure*}

To further assess allosteric binding from Boltz2, we selected a representative subset of 10 compounds across four Boltz2-estimated binding affinity ranges and evaluated their accuracy using an orthogonal physics-based method: absolute binding free energy (ABFE) \citep{chen2023enhancing}. Unlike relative binding free energy methods, which are optimal for congeneric series \citep{passaro2025boltz}, ABFE provides a rigorous and transferable measure of binding strength by estimating the free energy difference between bound and unbound states of a ligand. In this work we use absolute free energy perturbation (AFEP) formalism as implemented in Schrödinger FEP+ package. As shown in \autoref{fig:AFEP_validation}, Boltz2 estimated affinities correlate with AFEP derived binding affinities. Binned affinities show a significant variance across means as tested by a one-way ANOVA (p=0.0012), and non-binned affinities report a Pearson correlation of 0.7. In the absence of experimental data, these results indicate strong alignment between Boltz2 and AFEP, providing additional confidence in the use of this approach to search for new protein pockets and optimizing chemical matter binding to them.

\section{Conclusion} 

In this work, we revisited the REINFORCE algorithm, investigating additional extensions stemming from RL theory for de novo molecule generation with chemical language models. We investigated extensions believed to improve the validity, effectiveness and efficiency of learning, and exploration. Moreover, we uncovered a new reward-shaping mechanism to provide more control over the trade-off between optimization efficiency and prior policy regularization. Re-building REINFORCE by combining the investigated extensions demonstrates capacity for improvement of this algorithm demonstrated on the molecular optimization benchmark: showing state-of-the-art performance according to the author proposed metric of AUC Top-10, as well as excellent performance on our suite of metrics that additionally include chemistry. This performance is measured on a benchmark that focuses on exploitation, however, when tested on a case study to find putative allosteric JNK3 ligands, we also observed strong performance. Our version of the REINFORCE algorithm vastly outperforms the baseline model SynFlowNet in our sample efficiency scenario, and is able to optimize Boltz2 binding affinity considerably while maintaining drug-like property ranges of de novo molecules. Principally, we hope the results here act as a guide for implementing RL with CLMs. We make all the RL extensions, scripts for hyperparameter optimization, and the other CLMs mentioned above available in ACEGEN.

\subsection{Data and Software Availability}
All software used in this manuscript is freely available open-source under an MIT license. The RL configurations tested and used are incorporated into ACEGEN, available on 
GitHub (\url{https://github.com/Acellera/acegen-open}). The parameters of the pre-trained model used in this work are also available in ACEGEN repository. Benchmarking was conducted using MolScore which is available on GitHub (\url{https://github.com/MorganCThomas/MolScore}) and in the Python Package Index (\url{https://pypi.org/project/MolScore/}).

\subsection{Supporting Information}
The supporting information file includes supplementary methods, figures, and tables referenced in the text.

\subsection{Author Contributions}
\chadded{MT and AB experimented with and implemented the RL components. MT and JCGT conducted the experiment and analysis of the JNK3 case study using Boltz-2, and JCGT conducted the AFEP simulations. MT conducted the remaining analysis and wrote the manuscript under the supervision of GT, MA , JCGT, and GdF.}

\subsection{Acknowledgments}
This work was funded in part by the Flanders innovation \& entrepreneurship (VLAIO) project HBC.2021.1123

\subsection{Competing Interests}
\chadded{The authors declare no competing financial interests.}

\clearpage
\bibliography{references}

\end{document}


\tableofcontents

\newpage
\appendix
\onecolumn
\section{Reward shaping landscape}
\label{sec:reward_landscape}

\begin{figure}[ht]
\begin{center}
\centerline{\includegraphics[width=\textwidth]{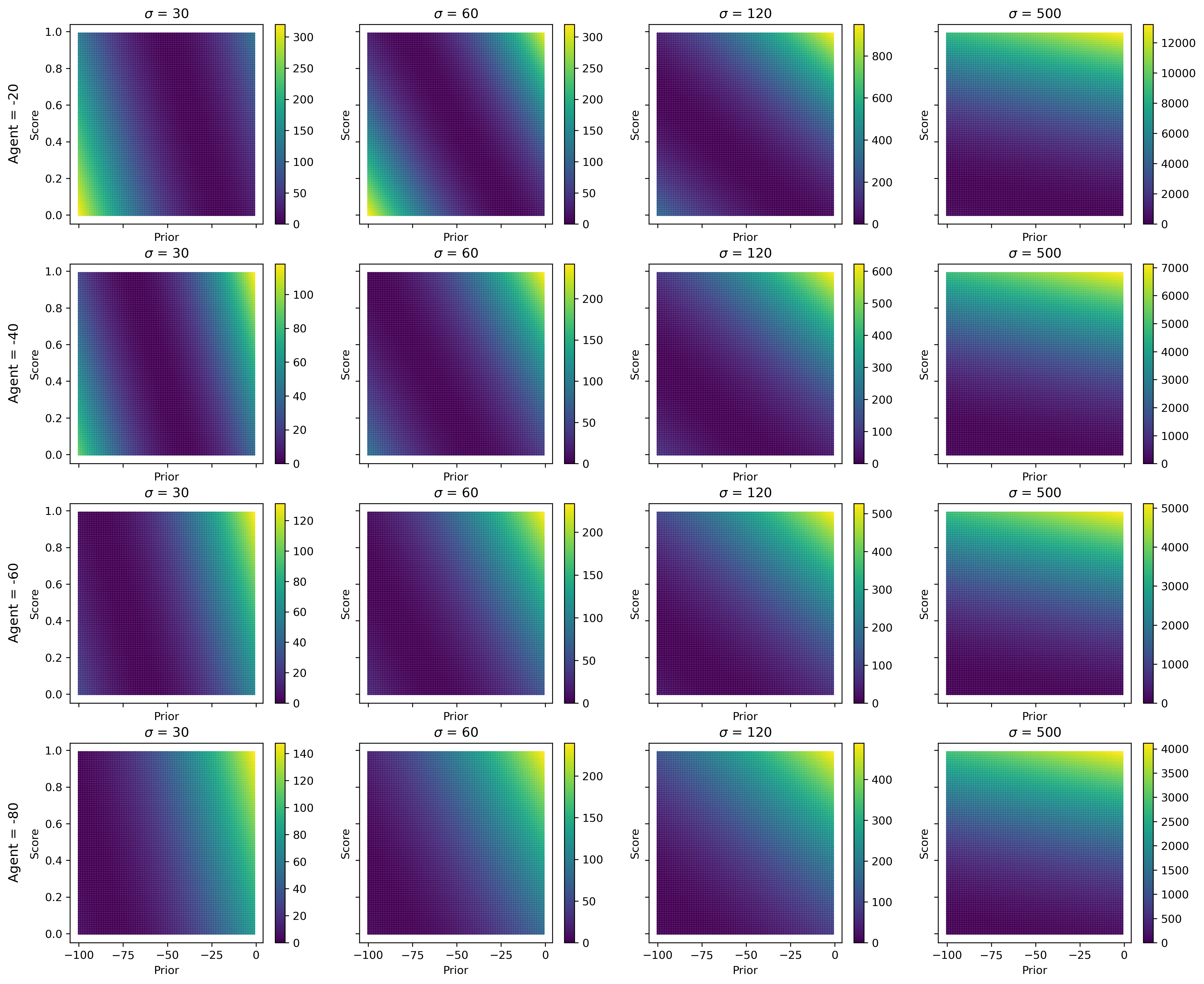}}
\caption{REINVENT reward shaping landscape depending on input reward (Score), Agent log-likelihood (Agent), Prior log-likelihood (Prior) and sigma ($\sigma$). Color reflects the reshaped reward.}
\label{fig:reinvent_reward_visualization}
\end{center}
\end{figure}

\begin{figure}[ht]
\begin{center}
\centerline{\includegraphics[width=\textwidth]{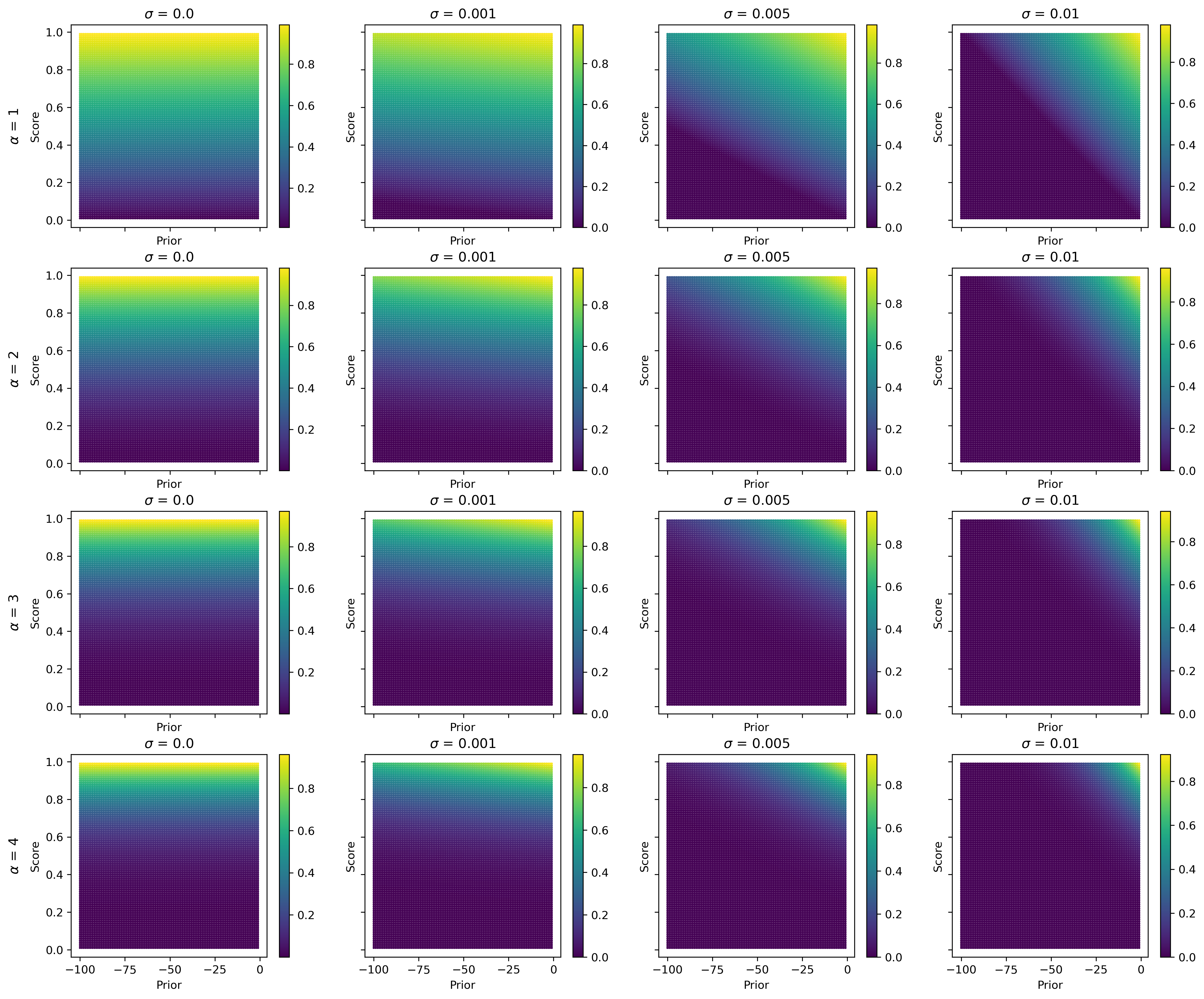}}
\caption{ACEGEN reward shaping landscape depending on input reward (Score), alpha ($\alpha$), Prior log-likelihood (Prior) and sigma ($\sigma$). Color reflects the reshaped reward.}
\label{fig:acegen_reward_visualization}
\end{center}
\end{figure}

\begin{figure}[ht]
\begin{center}
\centerline{\includegraphics[width=\textwidth]{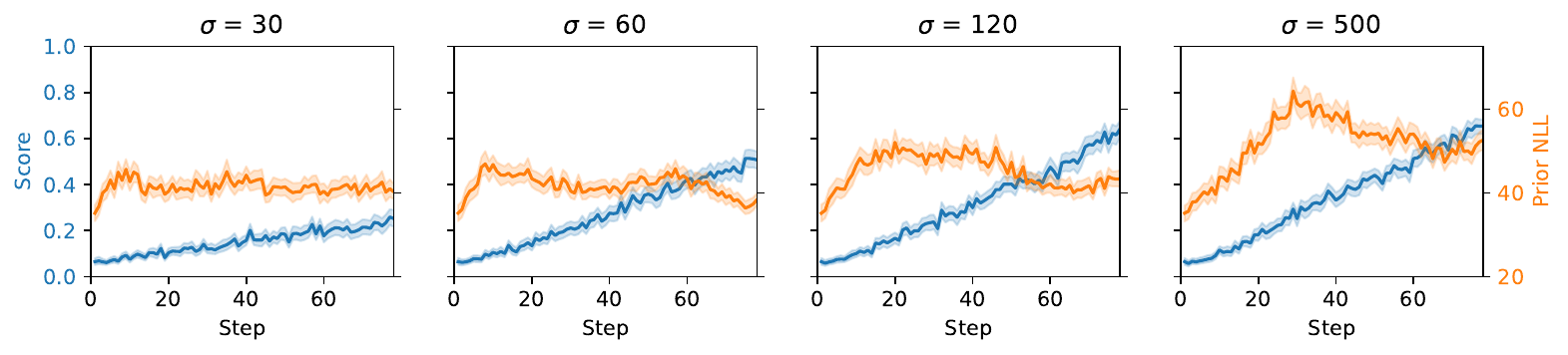}}
\caption{Performance of REINVENT at different values of $\sigma$ on the JNK3 MolOpt benchmark task. On the left y-axis (blue) is the JNK3 score during training step, while on the right y-axis (orange) is the negative log-likelihood (NLL) (lower is more likely, and hence better) of molecules according to the prior policy. Variables are measured during RL training steps until 10,000 molecules have been evaluated.}
\label{fig:reinvent_sigma_trade_off}
\end{center}
\end{figure}


\clearpage
\section{Baseline algorithm hyperparameters}
\label{sec:algorithm_hp}


\begin{table}[h!]
    \centering
    \caption{Hyperparameters for REINFORCE.}
    \vspace{0.2cm}
    \begin{tabular}{lc}
    \multicolumn{1}{c}{Hyperparameter} & \multicolumn{1}{c}{Value} \\ \toprule
    num\_envs & 128 \\ 
    total\_smiles & 10,000 \\ 
    model & GRU (embedding of size 256 + 3 layer GRU of size 512 + MLP) \\
    lr & 0.0001 \\
    experience\_replay & True \\
    replay\_buffer\_size & 100 \\
    replay\_batch\_size & 10 \\
    \end{tabular}
    \label{table:reinforce_hp}
\end{table}

\begin{table}[h!]
    \centering
    \caption{Hyperparameters for REINVENT.}
    \vspace{0.2cm}
    \begin{tabular}{lc}
    \multicolumn{1}{c}{Hyperparameter} & \multicolumn{1}{c}{Value} \\ \toprule
    num\_envs & 128 \\ 
    total\_smiles & 10,000 \\ 
    model & GRU (embedding of size 256 + 3 layer GRU of size 512 + MLP) \\
    lr & 0.0001 \\
    experience\_replay & True \\
    replay\_buffer\_size & 100 \\
    replay\_batch\_size & 10 \\
    sigma & 120 \\
    \end{tabular}
    \label{table:reinvent_hp}
\end{table}

\begin{table}[h!]
    \centering
    \caption{Hyperparameters for REINVENT$_{MolOpt}$.}
    \vspace{0.2cm}
    \begin{tabular}{lc}
    \multicolumn{1}{c}{Hyperparameter} & \multicolumn{1}{c}{Value} \\ \toprule
    num\_envs & 64 \\ 
    total\_smiles & 10,000 \\ 
    model & GRU (embedding of size 256 + 3 layer GRU of size 512 + MLP) \\
    lr & 0.0005 \\
    experience\_replay & True \\
    replay\_buffer\_size & 100 \\
    replay\_batch\_size & 24 \\
    sigma & 500 \\
    \end{tabular}
    \label{table:reinvent_molopt_hp}
\end{table}

\begin{table}[h!]
    \centering
    \caption{Hyperparameters for AHC.}
    \vspace{0.2cm}
    \begin{tabular}{lc}
    \multicolumn{1}{c}{Hyperparameter} & \multicolumn{1}{c}{Value} \\ \toprule
    num\_envs & 128 \\ 
    total\_smiles & 10,000 \\ 
    model & GRU (embedding of size 256 + 3 layer GRU of size 512 + MLP) \\
    lr & 0.0001 \\
    experience\_replay & True \\
    replay\_buffer\_size & 100 \\
    replay\_batch\_size & 10 \\
    sigma & 60 \\
    topk & 0.5 \\
    \end{tabular}
    \label{table:AHC_hp}
\end{table}

\begin{table}[H]
    \centering
    \caption{Hyperparameters for ACEGEN$_{Practical}$.}
    \vspace{0.2cm}
    \begin{tabular}{lc}
    \multicolumn{1}{c}{Hyperparameter} & \multicolumn{1}{c}{Value} \\ \toprule
    num\_envs & 128 \\ 
    total\_smiles & 10,000 \\ 
    model & GRU (embedding of size 256 + 3 layer GRU of size 512 + MLP) \\
    lr & 0.0001 \\
    experience\_replay & True \\
    replay\_batch\_size & 10 \\
    replay\_buffer\_size & 100 \\
    replay\_sampler & uniform \\
    sigma & 0.005 \\
    topk & 0.5 \\
    alpha & 5 \\
    baseline & mab \\
    \end{tabular}
    \label{table:acegen_hp_pract}
\end{table}

\begin{table}[H]
    \centering
    \caption{Hyperparameters for ACEGEN$_{MolOpt}$.}
    \vspace{0.2cm}
    \begin{tabular}{lc}
    \multicolumn{1}{c}{Hyperparameter} & \multicolumn{1}{c}{Value} \\ \toprule
    num\_envs & 32 \\ 
    total\_smiles & 10,000 \\ 
    model & GRU (embedding of size 256 + 3 layer GRU of size 512 + MLP) \\
    lr & 0.0001 \\
    experience\_replay & True \\
    replay\_batch\_size & 50 \\
    replay\_buffer\_size & 100 \\
    replay\_sampler & prioritized \\
    sigma & 0.001 \\
    topk & 0.5 \\
    alpha & 3 \\
    baseline & False \\
    \end{tabular}
    \label{table:acegen_hp_molopt}
\end{table}


\clearpage
\section{ACEGEN hyperparameter optimization}
\label{sec:hp_optimization}

\begin{table}[H]
    \centering
    \caption{Hyperparameter search space defined for ACEGEN.}
    \vspace{0.2cm}
    \begin{tabular}{lc}
    \multicolumn{1}{c}{Hyperparameter} & \multicolumn{1}{c}{Value} \\ \toprule
    num\_envs & 32, 64, 128, 256 \\ 
    total\_smiles & 10,000 \\ 
    model & GRU (embedding of size 256 + 3 layer GRU of size 512 + MLP) \\
    lr & 0.0001, 0.0005 \\
    lr\_annealing & False, True \\
    experience\_replay & False, True \\
    replay\_batch\_size & 10, 20, 50 \\
    replay\_buffer\_size & 100, 500 \\
    replay\_sampler & uniform, prioritized \\
    sigma & 0, 1e-3, 5e-3, 1e-2 \\
    kl\_coef & 0, 5e-3, 1e-2, 5e-2 \\
    alpha & 1, 2, 3, 4, 5 \\
    topk & 0.25, 0.5, 0.75, 1.0 \\
    baseline & False, mab, loo \\
    entropy\_coef & 0, 1e-3, 1e-2 \\
    likely\_penalty & 0, 10, 50 \\
    rnd\_coef & 0, 0.5, 1.0 \\
    \end{tabular}
    \label{table:acegen_hp_sweep}
\end{table}

\begin{figure}[ht]
\begin{center}
\centerline{\includegraphics[width=\textwidth]{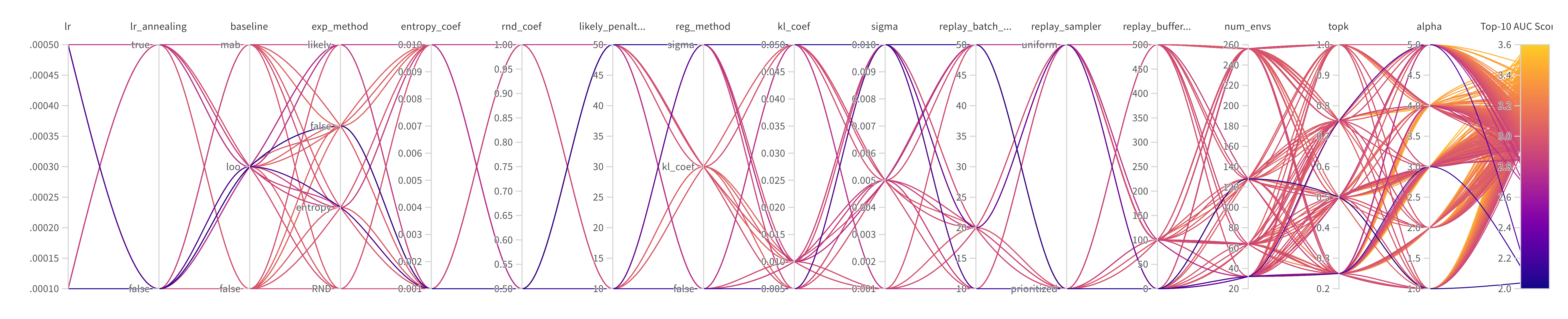}}
\caption{Sweep plot visualizing the effect of hyperparameters on the performance of "Osimertinib MPO" and "Median molecules 2" as measured by Top-10 AUC.}
\label{fig:hp_sweep_visualization}
\end{center}
\end{figure}


\clearpage
\section{Performance of individual REINFORCE extensions}
\label{sec:reinforce_extensions}

\begin{table*}[h!]
\small
\caption{Effect of different reward baselines including a moving average baseline (MAB) and leave-one-out baseline (LOO) on MolOpt benchmark performance.}
\label{tab:baselines}
\begin{tabular}{l|ccc}
 & REINFORCE & MAB & LOO \\
\hline
Valid & 21.74 ± 0.04 & \textbf{21.94 ± 0.02} & 21.92 ± 0.02 \\
Top-10 Avg & 14.74 ± 0.09 & 15.63 ± 0.11 & \textbf{15.69 ± 0.10} \\
Top-10 AUC & 13.21 ± 0.07 & \textbf{13.76 ± 0.07} & 13.76 ± 0.09 \\
Unique & \textbf{22.35 ± 0.05} & 21.66 ± 0.10 & 21.40 ± 0.11 \\
\hdashline
B\&T-CF & \textbf{14.15 ± 0.23} & 12.90 ± 0.15 & 12.60 ± 0.17 \\
B\&T-CF Top-10 Avg (Div) & 13.81 ± 0.09 & \textbf{14.68 ± 0.10} & 14.67 ± 0.13 \\
B\&T-CF Top-10 AUC (Div) & 12.51 ± 0.07 & \textbf{12.97 ± 0.06} & 12.89 ± 0.07 \\
B\&T-CF Diversity (SEDiv@1k) & \textbf{17.61 ± 0.18} & 15.49 ± 0.15 & 15.57 ± 0.12 \\
\end{tabular}
\end{table*}

\vspace*{0.5 cm}

\begin{table*}[h!]
\small
\caption{Effect of different learning rates including cosine annealing back to REINFORCE default ($\rightarrow$) on MolOpt benchmark performance.}
\label{tab:learning_rates}
\resizebox{\textwidth}{!}{%
\begin{tabular}{l|ccccc}
 & REINFORCE (1e-4) & 5e-4 & 5e-4 $\rightarrow$ 1e-4 & 1e-3 & 1e-3 $\rightarrow$ 1e-4 \\
\hline
Valid & \textbf{21.74 ± 0.04} & 21.36 ± 0.08 & 21.61 ± 0.07 & 20.71 ± 0.25 & 21.42 ± 0.18 \\
Top-10 Avg & 14.74 ± 0.09 & \textbf{14.93 ± 0.18} & 14.86 ± 0.18 & 14.33 ± 0.18 & 14.36 ± 0.27 \\
Top-10 AUC & 13.21 ± 0.07 & \textbf{13.83 ± 0.14} & 13.73 ± 0.14 & 13.54 ± 0.14 & 13.43 ± 0.22 \\
Unique & \textbf{22.35 ± 0.05} & 18.80 ± 0.27 & 19.92 ± 0.38 & 17.62 ± 0.29 & 16.59 ± 0.44 \\
\hdashline
B\&T-CF & \textbf{14.15 ± 0.23} & 7.36 ± 0.39 & 9.85 ± 0.52 & 3.92 ± 0.31 & 4.83 ± 0.44 \\
B\&T-CF Top-10 Avg (Div) & 13.81 ± 0.09 & \textbf{13.98 ± 0.16} & 13.92 ± 0.16 & 13.18 ± 0.18 & 13.29 ± 0.26 \\
B\&T-CF Top-10 AUC (Div) & 12.51 ± 0.07 & \textbf{12.97 ± 0.12} & 12.87 ± 0.12 & 12.52 ± 0.14 & 12.44 ± 0.23 \\
B\&T-CF Diversity (SEDiv@1k) & \textbf{17.61 ± 0.18} & 11.02 ± 0.38 & 11.11 ± 0.43 & 8.30 ± 0.37 & 8.19 ± 0.42 \\
\end{tabular}}
\end{table*}

\vspace*{0.5 cm}

\begin{table*}[h!]
\small
\caption{Effect of different top-k ratios on MolOpt benchmark performance.}
\label{tab:tok_k}
\begin{tabular}{l|cccc}
 & REINFORCE & 0.75 & 0.5 & 0.25 \\
\hline
Valid & \textbf{21.74 ± 0.04} & 21.52 ± 0.03 & 21.25 ± 0.05 & 21.06 ± 0.06 \\
Top-10 Avg & 14.74 ± 0.09 & 15.04 ± 0.13 & 15.69 ± 0.09 & \textbf{16.18 ± 0.07} \\
Top-10 AUC & 13.21 ± 0.07 & 13.38 ± 0.07 & 13.65 ± 0.07 & \textbf{14.03 ± 0.06} \\
Unique & \textbf{22.35 ± 0.05} & 22.15 ± 0.09 & 21.80 ± 0.11 & 21.39 ± 0.09 \\
\hdashline
B\&T-CF & \textbf{14.15 ± 0.23} & 12.77 ± 0.26 & 12.10 ± 0.21 & 11.66 ± 0.19 \\
B\&T-CF Top-10 Avg (Div) & 13.81 ± 0.09 & 14.05 ± 0.12 & 14.57 ± 0.11 & \textbf{15.06 ± 0.08} \\
B\&T-CF Top-10 AUC (Div) & 12.51 ± 0.07 & 12.61 ± 0.06 & 12.79 ± 0.06 & \textbf{13.13 ± 0.06} \\
B\&T-CF Diversity (SEDiv@1k) & \textbf{17.61 ± 0.18} & 17.59 ± 0.14 & 16.95 ± 0.15 & 15.89 ± 0.14 \\
\end{tabular}
\end{table*}

\vspace*{0.5 cm}

\begin{table*}[h!]
\caption{Effect of different experience replay parameters on MolOpt benchmark performance. The initial letter indicates the sampling type: prioritized proportional to the molecule’s reward (P) or uniform (U). The first number after the letter represents the replay batch size, and the final number indicates the experience replay buffer size.}
\label{tab:exp_replay}
\resizebox{\textwidth}{!}{%
\begin{tabular}{l|ccccccccc}
 & REINFORCE & +P10:100 & +U10:100 & +P20:100 & +U20:100 & +P10:500 & +U10:500 & +P20:500 & +U20:500 \\
\hline
Valid & 21.74 ± 0.04 & \textbf{21.77 ± 0.03} & 21.75 ± 0.03 & 21.74 ± 0.03 & 21.72 ± 0.03 & 21.77 ± 0.03 & 21.75 ± 0.03 & 21.77 ± 0.03 & 21.75 ± 0.03 \\
Top-10 Avg & 14.74 ± 0.09 & 15.85 ± 0.10 & 15.83 ± 0.06 & \textbf{16.28 ± 0.09} & 16.21 ± 0.10 & 15.21 ± 0.11 & 15.20 ± 0.10 & 15.49 ± 0.14 & 15.33 ± 0.11 \\
Top-10 AUC & 13.21 ± 0.07 & 13.67 ± 0.07 & 13.68 ± 0.06 & \textbf{14.04 ± 0.08} & 13.85 ± 0.12 & 13.38 ± 0.08 & 13.35 ± 0.07 & 13.47 ± 0.10 & 13.41 ± 0.09 \\
Unique & 22.35 ± 0.05 & 22.28 ± 0.10 & 22.29 ± 0.08 & 21.85 ± 0.12 & 22.01 ± 0.08 & 22.41 ± 0.06 & 22.45 ± 0.04 & 22.38 ± 0.07 & \textbf{22.49 ± 0.03} \\
\hdashline
B\&T-CF & 14.15 ± 0.23 & 14.34 ± 0.18 & 14.30 ± 0.18 & 13.83 ± 0.20 & 14.14 ± 0.18 & 14.64 ± 0.16 & 14.55 ± 0.19 & 14.69 ± 0.15 & \textbf{14.85 ± 0.18} \\
B\&T-CF Top-10 Avg (Div) & 13.81 ± 0.09 & 14.95 ± 0.12 & 14.89 ± 0.08 & \textbf{15.26 ± 0.13} & 15.22 ± 0.16 & 14.26 ± 0.09 & 14.33 ± 0.13 & 14.62 ± 0.14 & 14.42 ± 0.13 \\
B\&T-CF Top-10 AUC (Div) & 12.51 ± 0.07 & 12.91 ± 0.06 & 12.92 ± 0.05 & \textbf{13.19 ± 0.07} & 13.04 ± 0.15 & 12.67 ± 0.08 & 12.62 ± 0.07 & 12.74 ± 0.08 & 12.65 ± 0.07 \\
B\&T-CF Diversity (SEDiv@1k) & 17.61 ± 0.18 & 17.39 ± 0.15 & 17.43 ± 0.13 & 17.04 ± 0.13 & 17.15 ± 0.16 & 17.72 ± 0.13 & \textbf{17.82 ± 0.11} & 17.62 ± 0.15 & 17.79 ± 0.13 \\
\end{tabular}}
\end{table*}

\vspace*{0.5 cm}

\begin{table*}[h!]
\small
\caption{Effect of different $\alpha$ exponent values on MolOpt benchmark performance.}
\label{tab:alpha}
\resizebox{\textwidth}{!}{%
\begin{tabular}{l|cccccc}
 & REINFORCE & 2 & 3 & 4 & 5 & 6 \\
\hline
Valid & \textbf{21.74 ± 0.04} & 21.61 ± 0.03 & 21.55 ± 0.04 & 21.53 ± 0.04 & 21.52 ± 0.04 & 21.45 ± 0.03 \\
Top-10 Avg & 14.74 ± 0.09 & 15.66 ± 0.17 & 16.04 ± 0.09 & 16.39 ± 0.13 & \textbf{16.56 ± 0.13} & 16.54 ± 0.11 \\
Top-10 AUC & 13.21 ± 0.07 & 13.73 ± 0.11 & 14.02 ± 0.08 & 14.29 ± 0.08 & \textbf{14.38 ± 0.08} & 14.32 ± 0.08 \\
Unique & \textbf{22.35 ± 0.05} & 21.80 ± 0.12 & 21.28 ± 0.13 & 20.71 ± 0.17 & 20.24 ± 0.16 & 20.29 ± 0.16 \\
\hdashline
B\&T-CF & \textbf{14.15 ± 0.23} & 12.75 ± 0.19 & 12.34 ± 0.17 & 11.95 ± 0.22 & 11.62 ± 0.19 & 11.65 ± 0.21 \\
B\&T-CF Top-10 Avg (Div) & 13.81 ± 0.09 & 14.64 ± 0.17 & 15.00 ± 0.16 & 15.33 ± 0.11 & \textbf{15.41 ± 0.16} & 15.40 ± 0.12 \\
B\&T-CF Top-10 AUC (Div) & 12.51 ± 0.07 & 12.87 ± 0.08 & 13.12 ± 0.10 & 13.39 ± 0.09 & \textbf{13.45 ± 0.09} & 13.39 ± 0.08 \\
B\&T-CF Diversity (SEDiv@1k) & \textbf{17.61 ± 0.18} & 17.28 ± 0.14 & 17.04 ± 0.14 & 16.93 ± 0.13 & 16.86 ± 0.17 & 17.03 ± 0.16 \\
\end{tabular}}
\end{table*}

\vspace*{0.5 cm}

\begin{table*}[h!]
\small
\caption{Effect of different regularization strategies on MolOpt benchmark performance. Including different $\sigma$ values for the proposed reward shaping and different $\lambda_{KL}$ coefficients for regularization by KL divergence.}
\label{tab:reg}
\resizebox{\textwidth}{!}{
\begin{tabular}{l|ccccccc}
 & REINFORCE & $\sigma$=1e-3 & $\sigma$=5e-3 & $\sigma$=1e-2 & $\lambda_{KL}$=5e-3 & $\lambda_{KL}$=1e-2 & $\lambda_{KL}$=5e-2 \\
\hline
Valid & 21.74 ± 0.04 & 21.97 ± 0.03 & 22.32 ± 0.02 & \textbf{22.42 ± 0.01} & 21.81 ± 0.02 & 21.84 ± 0.02 & 21.89 ± 0.01 \\
Top-10 Avg & 14.74 ± 0.09 & 14.86 ± 0.11 & \textbf{14.87 ± 0.19} & 14.06 ± 0.16 & 14.84 ± 0.10 & 14.54 ± 0.10 & 13.94 ± 0.10 \\
Top-10 AUC & 13.21 ± 0.07 & \textbf{13.30 ± 0.08} & 13.23 ± 0.10 & 12.65 ± 0.09 & 13.20 ± 0.07 & 13.04 ± 0.07 & 12.67 ± 0.10 \\
Unique & 22.35 ± 0.05 & 21.98 ± 0.09 & 20.71 ± 0.10 & 20.34 ± 0.11 & 22.48 ± 0.04 & 22.60 ± 0.04 & \textbf{22.93 ± 0.05} \\
\hdashline
B\&T-CF & 14.15 ± 0.23 & 14.44 ± 0.20 & 15.12 ± 0.12 & 16.01 ± 0.12 & 14.59 ± 0.17 & 15.36 ± 0.12 & \textbf{16.75 ± 0.06} \\
B\&T-CF Top-10 Avg (Div) & 13.81 ± 0.09 & 13.90 ± 0.09 & \textbf{13.94 ± 0.15} & 13.38 ± 0.16 & 13.92 ± 0.09 & 13.71 ± 0.10 & 13.28 ± 0.11 \\
B\&T-CF Top-10 AUC (Div) & 12.51 ± 0.07 & \textbf{12.55 ± 0.06} & 12.51 ± 0.07 & 12.09 ± 0.07 & 12.48 ± 0.06 & 12.35 ± 0.06 & 12.07 ± 0.09 \\
B\&T-CF Diversity (SEDiv@1k) & 17.61 ± 0.18 & 16.62 ± 0.14 & 14.50 ± 0.13 & 13.87 ± 0.14 & 18.11 ± 0.13 & 18.45 ± 0.12 & \textbf{19.63 ± 0.08} \\
\end{tabular}}
\end{table*}

\vspace*{0.5 cm}

\begin{table*}[h!]
\small
\caption{Effect of different exploration strategies on MolOpt benchmark performance. Including different entropy coefficients $\lambda_{ENT}$, agent likelihood penalty coefficients $\lambda_{ALL}$, diversity filters (DF), and random network distillation coefficients $\lambda_{RND}$.}
\label{tab:exp}
\resizebox{\textwidth}{!}{%
\begin{tabular}{l|ccccccccccc}
 & REINFORCE & $\lambda_{ENT}$=1e-3 & $\lambda_{ENT}$=1e-2 & $\lambda_{ENT}$=1e-1 & $\lambda_{ALL}$=10 & $\lambda_{ALL}$=50 & $\lambda_{ALL}$=100 & DF=Unique & DF=Similar & $\lambda_{RND}$=0.5 & $\lambda_{RND}$=1 \\
\hline
Valid & \textbf{21.74 ± 0.04} & 20.81 ± 0.06 & 18.46 ± 0.41 & 4.84 ± 0.12 & 20.99 ± 0.04 & 19.41 ± 0.11 & 18.35 ± 0.10 & 21.71 ± 0.03 & 21.74 ± 0.03 & 20.85 ± 0.11 & 20.63 ± 0.14 \\
Top-10 Avg & 14.74 ± 0.09 & 14.63 ± 0.14 & 13.46 ± 0.27 & 10.72 ± 0.13 & \textbf{14.77 ± 0.10} & 14.08 ± 0.16 & 13.38 ± 0.12 & 14.74 ± 0.12 & 14.61 ± 0.14 & 13.26 ± 0.14 & 12.90 ± 0.13 \\
Top-10 AUC & \textbf{13.21 ± 0.07} & 13.07 ± 0.15 & 12.44 ± 0.21 & 10.57 ± 0.13 & 13.16 ± 0.06 & 12.70 ± 0.13 & 12.28 ± 0.09 & 13.18 ± 0.07 & 13.15 ± 0.08 & 12.02 ± 0.10 & 11.86 ± 0.09 \\
Unique & 22.35 ± 0.05 & 22.46 ± 0.05 & 22.85 ± 0.12 & 22.96 ± 0.01 & 22.64 ± 0.04 & 22.98 ± 0.00 & \textbf{23.00 ± 0.00} & 22.62 ± 0.03 & 22.45 ± 0.07 & 22.98 ± 0.00 & 22.99 ± 0.00 \\
\hdashline
B\&T-CF & 14.15 ± 0.23 & 13.03 ± 0.24 & 10.71 ± 0.23 & 2.24 ± 0.04 & 13.19 ± 0.19 & 9.93 ± 0.20 & 7.61 ± 0.19 & 14.13 ± 0.22 & \textbf{14.18 ± 0.18} & 12.84 ± 0.26 & 12.34 ± 0.26 \\
B\&T-CF Top-10 Avg (Div) & 13.81 ± 0.09 & 13.69 ± 0.17 & 12.62 ± 0.30 & 10.13 ± 0.09 & 13.81 ± 0.10 & 12.90 ± 0.17 & 12.27 ± 0.12 & \textbf{13.85 ± 0.10} & 13.74 ± 0.11 & 12.41 ± 0.12 & 12.09 ± 0.11 \\
B\&T-CF Top-10 AUC (Div) & \textbf{12.51 ± 0.07} & 12.32 ± 0.19 & 11.72 ± 0.21 & 9.99 ± 0.09 & 12.39 ± 0.06 & 11.78 ± 0.14 & 11.41 ± 0.09 & 12.46 ± 0.06 & 12.42 ± 0.06 & 11.36 ± 0.09 & 11.22 ± 0.08 \\
B\&T-CF Diversity (SEDiv@1k) & 17.61 ± 0.18 & 17.34 ± 0.19 & 19.31 ± 0.10 & 4.64 ± 0.03 & 18.41 ± 0.13 & 18.96 ± 0.11 & 19.81 ± 0.08 & 17.86 ± 0.14 & 17.73 ± 0.14 & 20.13 ± 0.13 & \textbf{20.61 ± 0.14} \\
\end{tabular}}
\end{table*}

\begin{figure}[ht]
\begin{center}
\centerline{\includegraphics[width=\textwidth]{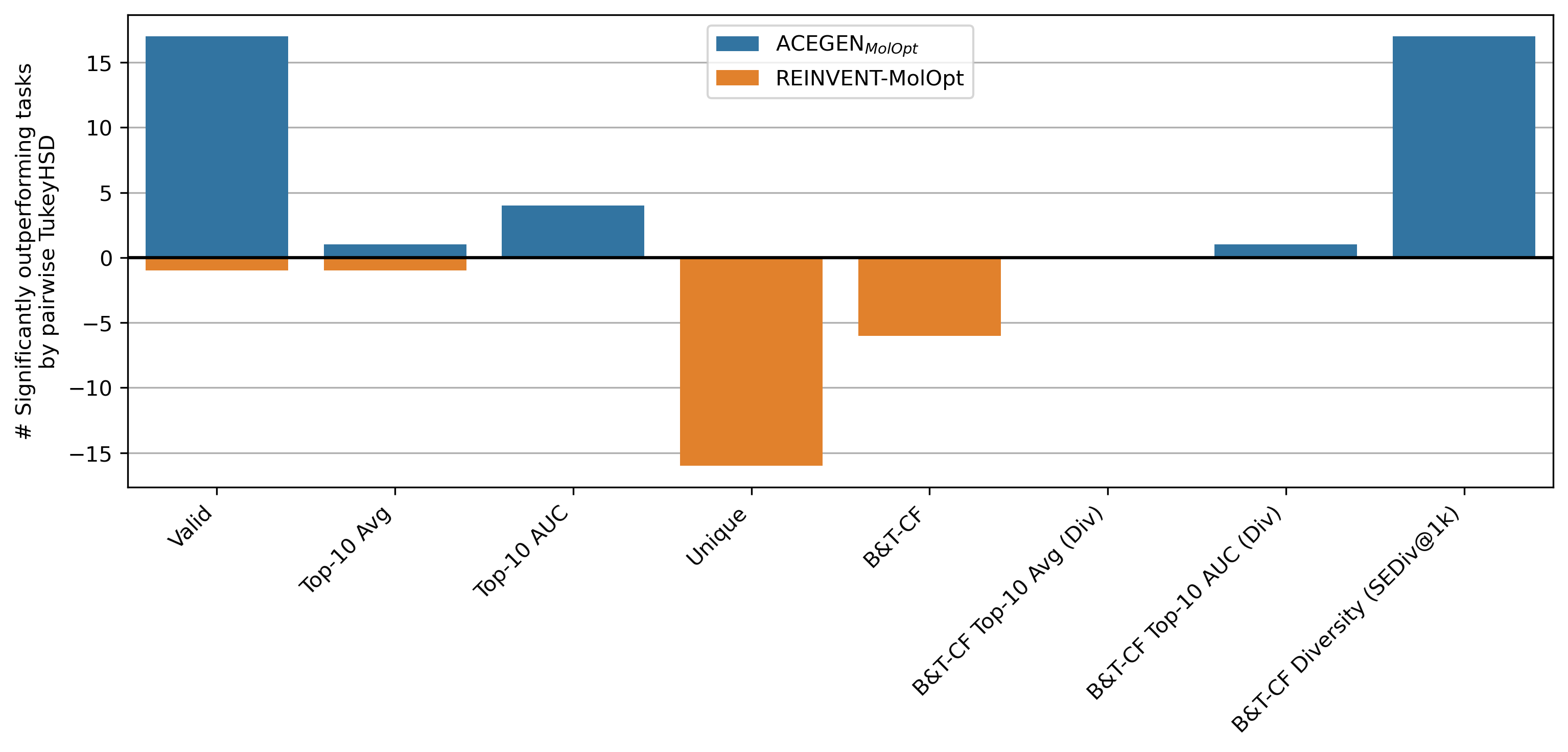}}
\caption{Per task significance performance comparison between ACEGEN$_{MolOpt}$ and REINVENT-MolOpt by pairwise TukeyHSD on key metrics. Positive count means ACEGEN$_{MolOpt}$ peformed better, negative count means REINVENT-MolOpt performed better. For example, ACEGEN$_{MolOpt}$ performs significantly better than REINVENT-MolOpt in Top-10 AUC in 4 out of 23 tasks. Note that although REINVENT-MolOpt generates significantly more unique molecules in 16 tasks, ACEGEN$_{MolOpt}$ generates significantly more diverse solutions in 16 tasks.}
\label{fig:per_task_significance}
\end{center}
\end{figure}

\clearpage
\section{Optimizing Boltz2 estimated binding affinity}
\label{sec:JNK3_boltz2}

\subsection{Boltz2 reward function}

To use Boltz2 \citep{passaro2025boltz} as a reward function, we integrated Boltz2 into MolScore \citep{thomas2023molscore} adding it to the suite of other scoring functions available. The yaml files to run Boltz2 co-folding and estimated binding affinity for ATP-competitive and allosteric can be found in Listing \autoref{lst:boltz_affinity} and \autoref{lst:boltz_allo_affinity}, respectively. The reward for molecules was calculated as the arithmetic mean of several properties ($p_i$) the estimated binding affinity, the estimated binding probability, and the QED \cite{bickerton2012quantifying} (in order to encourage molecules to maintain drug-like properties). The QED and and estimated binding probability are already in the range [0, 1], however, estimated affinity is in the range [2, -3]. In this case, estimated affinity values correspond to the log$_{10}$ scale of potency centered at 1 $\mu$M. Hence 2 has a potency of 100 $\mu$M and -3 has a potency of 1 nM. This is therefore transformed to a range [0,1] by maxmin normalization, where 1 is the best (lowest) value observed, and 0 is the worst (highest) value observed. The maximum and minimum values are updated throughout learning as new molecules are sampled. Lastly, to encourage diverse candidate solutions, we use a ScaffoldSimilarityECFP diversity filter (described elsewhere \citep{thomas2022augmented, thomas2023molscore}) to penalize the reward when already commonly observed molecular scaffolds and resampled. The full hyperparameters for this reward mechanism are available in Listing \autoref{lst:molscore_boltz} which can be passed to MolScore to exactly reproduce the drug design task (alongside the Boltz2 YAML files). 

\begin{equation}\label{eq:molscore_reward}
R(x) = \frac{1}{n} \sum_{i=1}^{n} p_i(x)
\end{equation}

\clearpage
\begin{center}
\begin{minted}{yaml}
version: 1  # Optional, defaults to 1
sequences:
  - protein:
      id: [A]
      sequence: LHFLYYCSEPTLDVKIAFCQGFDKQVDVSYIAKHYNMSKSKVDNQFYSVEVGDSTFTVLKRYQNLKPIGSGAQGIVCAAYDAVLDRNVAIKKLSRPFQNQTHAKRAYRELVLMKCVNHKNIISLLNVFTPQKTLEEFQDVYLVMELMDANLCQVIQMELDHERMSYLLYQMLCGIKHLHSAGIIHRDLKPSNIVVKSDCTLKILDFGLARTAGTSFMMTPYVVTRYYRAPEVILGMGYKENVDIWSVGCIMGEMVRHKILFPGRDYIDQWNKVIEQLGTPCPEFMKKLQPTVRNYVENRPKYAGLTFPKLFPDSLFPADSEHNKLKASQARDLLSKMLVIDPAKRISVDDALQHPYINVWYDPAEVEAPPPQIYDKQLDEREHTIEEWKELIYKEVMNSEEKTKNGVVKGQPSPSGAAVNSSESLPPSSSVNDISSMSTDQTLASDTDSSLEASAGPLGCCR
properties:
  - affinity:
    binder: Z
\end{minted}
\captionof{listing}{Boltz2 YAML configuration for ATP-competitive JNK3 co-folding.}
\label{lst:boltz_affinity}
\end{center}

\begin{center}
\begin{minted}{yaml}
version: 1  # Optional, defaults to 1
sequences:
  - protein:
      id: [A]
      sequence: LHFLYYCSEPTLDVKIAFCQGFDKQVDVSYIAKHYNMSKSKVDNQFYSVEVGDSTFTVLKRYQNLKPIGSGAQGIVCAAYDAVLDRNVAIKKLSRPFQNQTHAKRAYRELVLMKCVNHKNIISLLNVFTPQKTLEEFQDVYLVMELMDANLCQVIQMELDHERMSYLLYQMLCGIKHLHSAGIIHRDLKPSNIVVKSDCTLKILDFGLARTAGTSFMMTPYVVTRYYRAPEVILGMGYKENVDIWSVGCIMGEMVRHKILFPGRDYIDQWNKVIEQLGTPCPEFMKKLQPTVRNYVENRPKYAGLTFPKLFPDSLFPADSEHNKLKASQARDLLSKMLVIDPAKRISVDDALQHPYINVWYDPAEVEAPPPQIYDKQLDEREHTIEEWKELIYKEVMNSEEKTKNGVVKGQPSPSGAAVNSSESLPPSSSVNDISSMSTDQTLASDTDSSLEASAGPLGCCR
  - ligand:
      id: [B]
      smiles:  Nc1ncnc2c1ncn2[C@@H]1O[C@H](COP(=O)(O)OP(=O)(O)OP(=O)(O)O)[C@@H](O)[C@H]1O # ATP
properties:
  - affinity:
    binder: Z
\end{minted}
\captionof{listing}{Boltz2 YAML configuration for allosteric JNK3 co-folding with ATP stated to bind to the ATP site.}
\label{lst:boltz_allo_affinity}
\end{center}

\clearpage
\begin{center}
\begin{minted}{json}
{
  "task": "JNK3_binding",
  "output_dir": "./",
  "load_from_previous": false,
  "logging": false,
  "monitor_app": false,
  "termination_exit": false,
  "scoring_functions": [
    {
      "name": "Boltz",
      "run": true,
      "parameters": {
        "input_path": "/path/to/boltz_affinity.yaml",
        "prefix": "Boltz",
        "env_engine": "mamba",
        "cache": "~/.boltz",
        "devices": [
          "0",
          "1",
          "2",
          "3"
        ],
        "accelerator": "gpu",
        "recycling_steps": 3,
        "sampling_steps": 200,
        "diffusion_samples": 1,
        "step_scale": 1.638,
        "output_format": "mmcif",
        "num_workers": 2,
        "override": false,
        "use_msa_server": true,
        "msa_server_url": "https://api.colabfold.com",
        "msa_pairing_strategy": "greedy",
        "write_full_pae": false,
        "write_full_pde": false
      }
    },
    {
      "name": "MolecularDescriptors",
      "run": true,
      "parameters": {
        "prefix": "desc",
        "n_jobs": 1
      }
    }
  ],
  "scoring": {
    "metrics": [
      {
        "name": "Boltz_affinity_pred_value",
        "filter": false,
        "weight": 1.0,
        "modifier": "norm",
        "parameters": {
          "objective": "minimize"
        }
      },
      {
        "name": "Boltz_affinity_probability_binary",
        "filter": false,
        "weight": 1.0,
        "modifier": "raw",
        "parameters": {}
      },
      {
        "name": "desc_QED",
		"filter": false,
        "weight": 1.0,
        "modifier": "raw",
        "parameters": {}
      }
    ],
    "method": "amean"
  },
  "diversity_filter": {
    "run": true,
    "name": "ScaffoldSimilarityECFP",
    "parameters": {
      "nbmax": 50,
      "minscore": 0.5,
      "minsimilarity": 0.8,
      "radius": 2,
      "useFeatures": false,
      "bits": 1024,
      "outputmode": "linear"
    }
  }
}
\end{minted}
\captionof{listing}{MolScore configuration file for the JNK3 binding affinity task. This can be used to reproduce the same scoring/evaluation mechanism.}
\label{lst:molscore_boltz}
\end{center}

\subsection{SynFlowNet}

We re-implemented SynFlowNet \citep{cretu2024synflownet} with code forked from \url{https://github.com/mirunacrt/synflownet}. Minimal changes were made to the code to use MolScore \cite{thomas2023molscore} to calculate the reward for SMILES, ensuring equal objective design across algorithms. Note that MolScore returns a score in the range [0,1] whilst SynFlowNet expected a score in the range [0, 100], therefore scores were transformed to match the expected range in SynFlowNet, constituting the only notable code change. None of the default parameters were modified which are listed in Listing \autoref{lst:hp_sfn}. To ensure that SynFlowNet was working as expected, we ran additional experiments on two tasks reported in the original publication, GSK3$\beta$ and DRD2. \autoref{tab:sfn_val} shows similar performance as expected, in one case slightly worse, in the other case slightly better. In our case we ran GSK3$\beta$ and DRD2 as implemented in MolScore which may lead to small deviations, however, most importantly similar learning is observed within a budget of 10,000 molecule evaluations. 

\begin{center}
\begin{minted}{yaml}
desc: noDesc
log_dir: ./logs/debug_run_reactions_task_2025-07-08_00-20-43
resume: null
device: cuda
seed: 43
validate_every: 5000
checkpoint_every: null
store_all_checkpoints: false
print_every: 1
start_at_step: 0
num_final_gen_steps: 0
num_validation_gen_steps: 10
num_training_steps: 156
num_workers: 0
hostname: localhost
pickle_mp_messages: false
mp_buffer_size: 536870912
git_hash: 3df8f52
overwrite_existing_exp: true
algo:
  method: TB
  num_from_policy: 64
  num_from_dataset: 0
  num_from_buffer_for_pb: 64
  valid_num_from_policy: 64
  valid_num_from_dataset: 0
  max_len: 3
  max_nodes: 9
  max_edges: 128
  illegal_action_logreward: -75.0
  illegal_bck_traj_reward: -1.0
  bck_reward_exponent: 1.0
  synthesis_cost_as_bck_reward: false
  strict_forward_policy: false
  strict_bck_masking: false
  train_random_action_prob: 0.0
  train_det_after: null
  valid_random_action_prob: 0.0
  sampling_tau: 0.99
  tb:
    bootstrap_own_reward: false
    epsilon: null
    reward_loss_multiplier: 1.0
    variant: TB
    do_correct_idempotent: false
    do_parameterize_p_b: true
    do_predict_n: false
    do_sample_p_b: false
    do_length_normalize: false
    subtb_max_len: 128
    Z_learning_rate: 0.001
    Z_lr_decay: 50000.0
    cum_subtb: true
    loss_fn: MSE
    loss_fn_par: 1.0
    n_loss: none
    n_loss_multiplier: 1.0
    reinforce_loss_multiplier: 1.0
    bck_entropy_loss_multiplier: 1.0
    mle_loss_multiplier: 1.0
    backward_policy: MaxLikelihood
model:
  num_layers: 4
  num_emb: 128
  dropout: 0.0
  graph_transformer:
    num_heads: 2
    ln_type: pre
    num_mlp_layers: 0
    concat_heads: true
    continuous_action_embs: true
    fingerprint_type: morgan_1024
    fingerprint_path: null
opt:
  opt: adam
  learning_rate: 0.0001
  lr_decay: 2000.0
  weight_decay: 1.0e-08
  momentum: 0.9
  clip_grad_type: norm
  clip_grad_param: 10.0
  adam_eps: 1.0e-08
replay:
  use: false
  capacity: 10000
  warmup: 64
  hindsight_ratio: 0.0
  num_from_replay: 0
  num_new_samples: 64
task:
  reactions_task:
    templates_filename: hb.txt
    reverse_templates_filename: null
    reward: null # Calculated by MolScore
    building_blocks_filename: enamine_bbs.txt
    precomputed_bb_masks_filename: precomputed_bb_masks_enamine_bbs.pkl
    building_blocks_costs: null
    sanitize_building_blocks: false
cond:
  valid_sample_cond_info: true
  temperature:
    sample_dist: constant
    dist_params:
    - 32.0
    num_thermometer_dim: 32
  moo:
    num_objectives: 2
    num_thermometer_dim: 16
  weighted_prefs:
    preference_type: dirichlet
    preference_param: 1.5
  focus_region:
    focus_type: centered
    use_steer_thermomether: false
    focus_cosim: 0.98
    focus_limit_coef: 0.1
    focus_model_training_limits:
    - 0.25
    - 0.75
    focus_model_state_space_res: 30
    max_train_it: 20000
reward: null # Calculated by MolScore
\end{minted}
\captionof{listing}{Default hyperparameters used for SynFlowNet. Note that the reward parameter is ignored as this is calculated using MolScore instead.}
\label{lst:hp_sfn}
\end{center}

\begin{table*}[h!]
\centering
\caption{Using SynFlowNet in this work on two example tasks reported in the original publication \citep{cretu2024synflownet}. In our case, GSK3$\beta$ performs slightly worse, but DRD2 performs slightly better. Likewise we ran the tasks in 3 replicates.}
\label{tab:sfn_val}
\begin{tabular}{lcc}
  \toprule
  Task & SynFlowNet \citep{cretu2024synflownet} (reported) & SynFlowNet (our findings) \\ 
  \hline
  GSK3$\beta$ & 0.691 $\pm$ 0.034 & 0.567 $\pm$ 0.041 \\
  DRD2$\beta$ & 0.885 $\pm$ 0.027 & 0.927 $\pm$ 0.018 \\
  \bottomrule

\end{tabular}
\end{table*}

\subsection{De novo generated compounds}

\begin{figure}[ht]
\begin{center}
\centerline{\includegraphics[width=\textwidth]{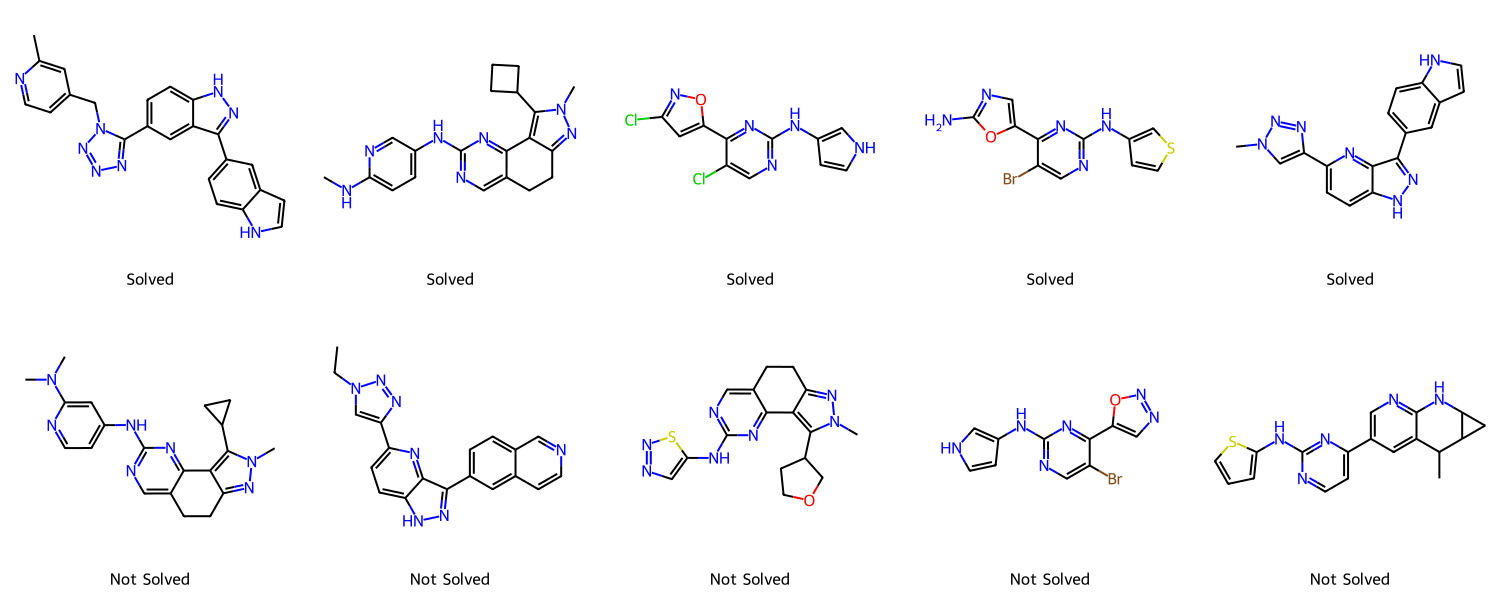}}
\caption{Examples of AiZynthFinder solved (top row) and unsolved (bottom row) ACEGEN-generated compounds from the top 100 predicted orthosteric binders (ATP-site).}
\label{fig:ATPc_synth_examples}
\end{center}
\end{figure}

\begin{figure}[ht]
\begin{center}
\centerline{\includegraphics[width=\textwidth]{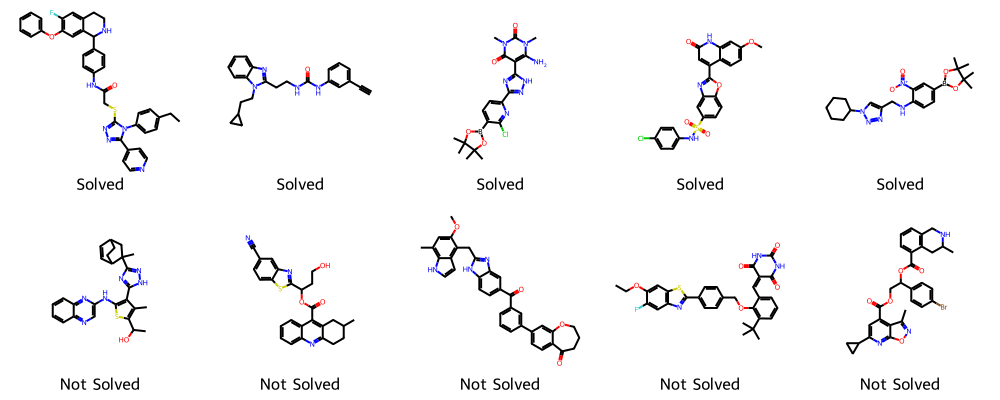}}
\caption{Examples of AiZynthFinder solved (top row) and unsolved (bottom row) SynFlowNet-generated compounds from the top 100 predicted orthosteric binders (ATP-site).}
\label{fig:SFN_synth_examples}
\end{center}
\end{figure}

\begin{figure}[ht]
\begin{center}
\centerline{\includegraphics[width=\textwidth]{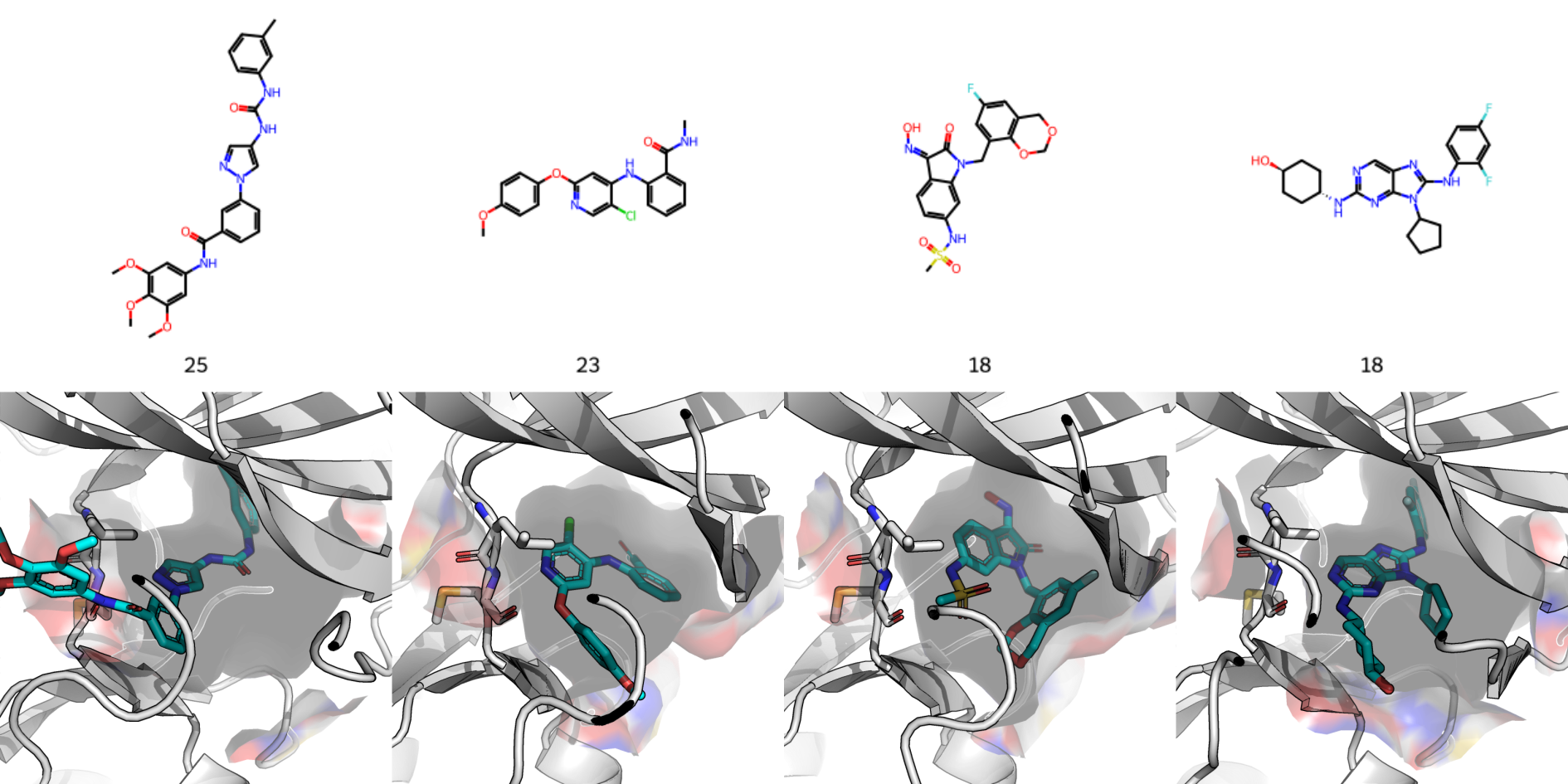}}
\caption{Examples of JNK3 ligands and their co-folded structure into the orthosteric pocket (ATP-site). Centroids of the 4 largest clusters are shown, alongside the number of cluster members. Molecules were clustered via Butina clustering of the ECFP4 fingerprints of their Bemis-Murcko scaffolds (using a distance threshold of 0.25)}
\label{fig:JNK3_ATPc_examples}
\end{center}
\end{figure}

\begin{figure}[ht]
\begin{center}
\centerline{\includegraphics[width=\textwidth]{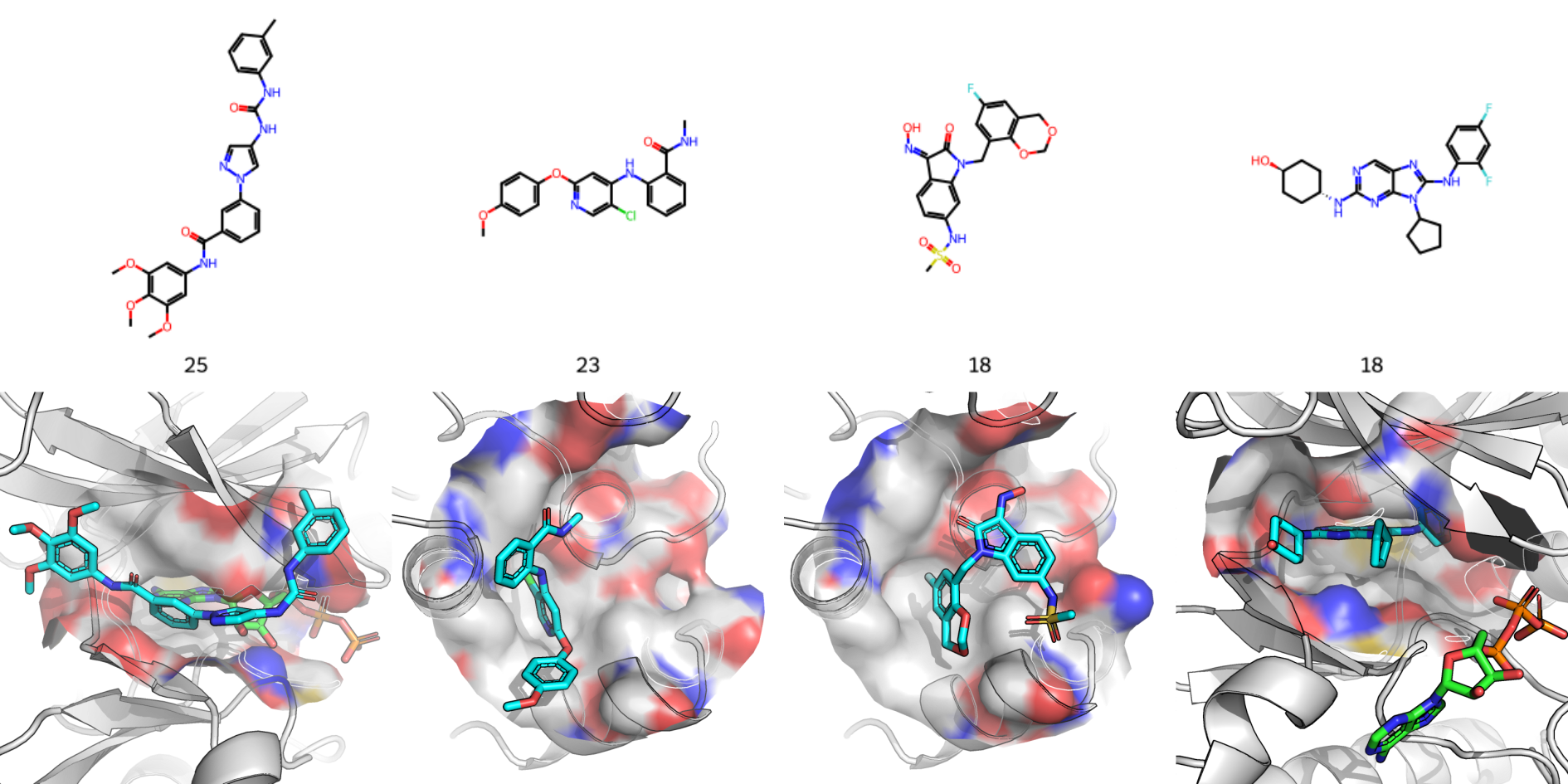}}
\caption{Examples of JNK3 ligands and their co-folded structure into an allosteric pocket (i.e., co-folded with ATP). Centroids of the 4 largest clusters are shown, alongside the number of cluster members. Molecules were clustered via Butina clustering of the ECFP4 fingerprints of their Bemis-Murcko scaffolds (using a distance threshold of 0.25)}
\label{fig:JNK3_nATPc_examples}
\end{center}
\end{figure}

\begin{figure}[ht]
\begin{center}
\centerline{\includegraphics[width=\textwidth]{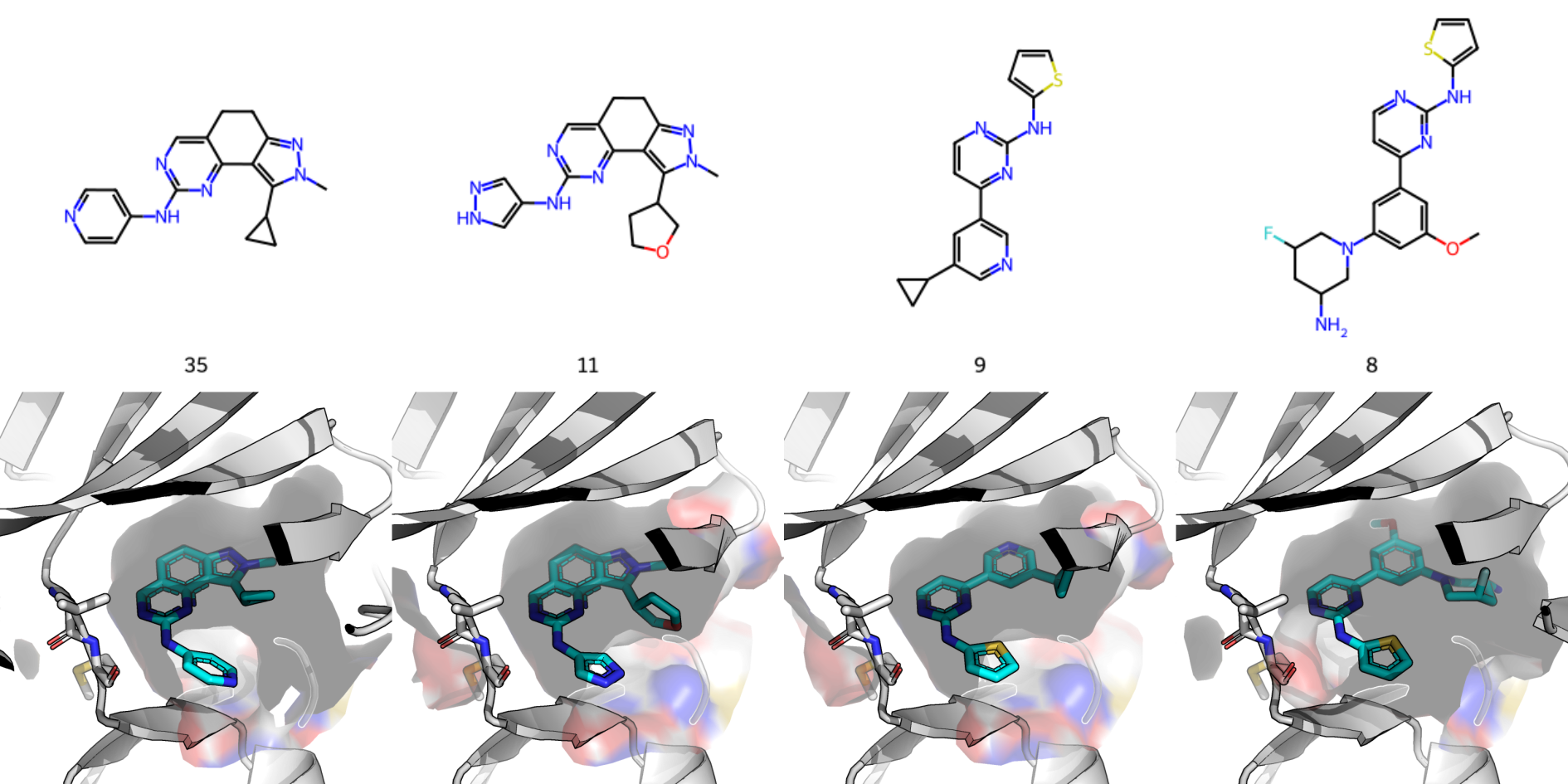}}
\caption{Examples of de novo compounds and their co-folded structure into the orthosteric pocket (ATP-site). Centroids of the 4 largest clusters are shown, alongside the number of cluster members. Molecules were clustered via Butina clustering of the ECFP4 fingerprints of their Bemis-Murcko scaffolds (using a distance threshold of 0.25)}
\label{fig:ATPc_examples}
\end{center}
\end{figure}

\begin{figure}[ht]
\begin{center}
\centerline{\includegraphics[width=\textwidth]{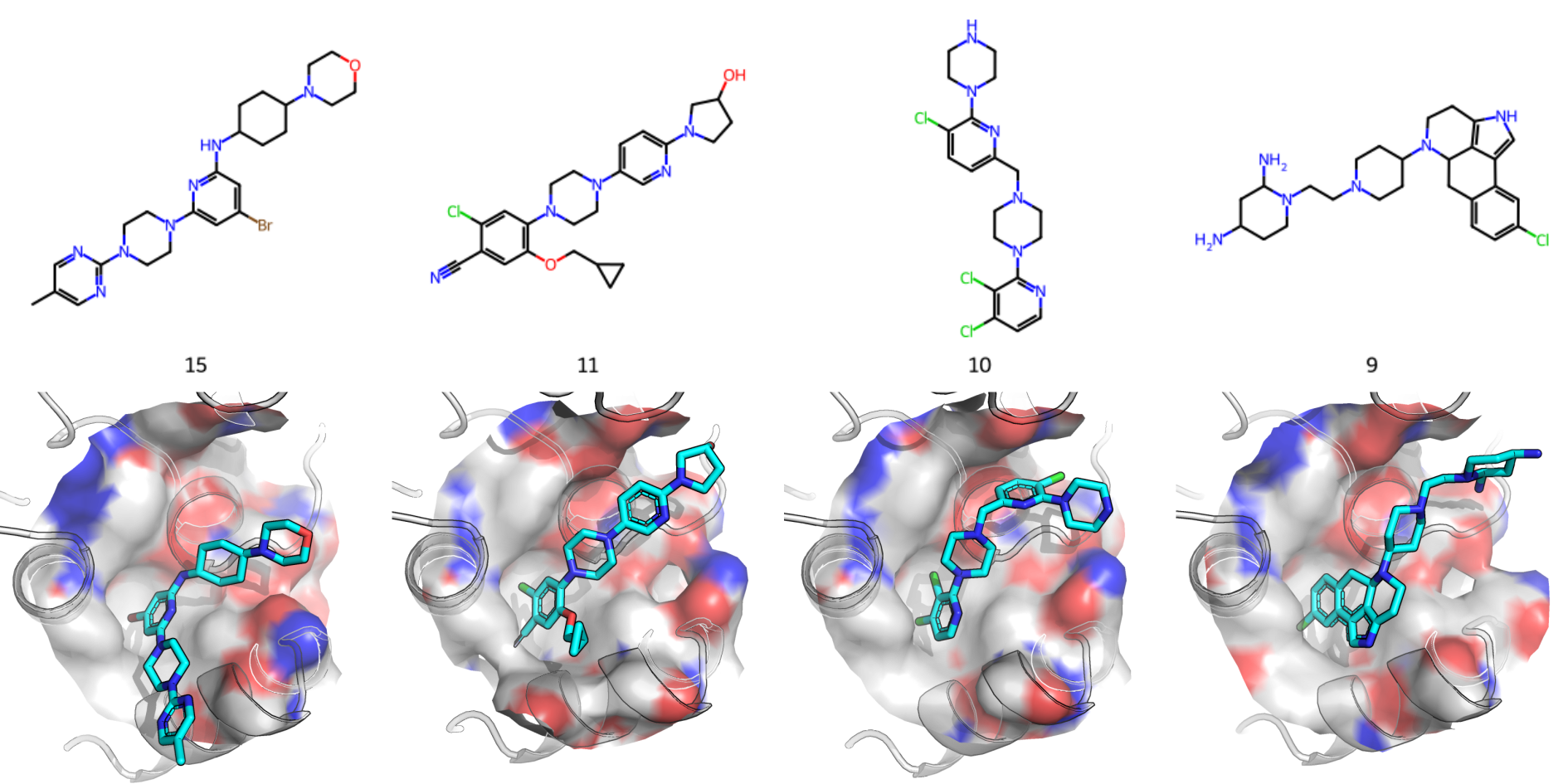}}
\caption{Examples of de novo compounds and their co-folded structure into an allosteric pocket (i.e., co-folded with ATP). Centroids of the 4 largest clusters are shown, alongside the number of cluster members. Molecules were clustered via Butina clustering of the ECFP4 fingerprints of their Bemis-Murcko scaffolds (using a distance threshold of 0.25)}
\label{fig:nATPc_examples}
\end{center}
\end{figure}

\clearpage

\subsection{Absolute binding free energy}

Preparation of the co-folded structures was done using Schrödinger protein preparation wizard \citep{madhavi2013protein}. This preparation step involves assigning bond orders, adding hydrogens, creating
disulfide bonds, generating most likely tautomer and protonation state, and a restrained minimization using OPLS4 forcefield. AFEP calculations were done using FEP+ suite from Schrödinger \citep{chen2023enhancing} with default parameters; chemical potential ensemble, 5 ns lambda windows, and 1 ns replica exchange solute tempering molecular dynamics for enhanced sampling. Both steps were performed using 2024-4 Schrödinger release.

\begin{figure}[ht]
\begin{center}
\centerline{\includegraphics[width=\textwidth]{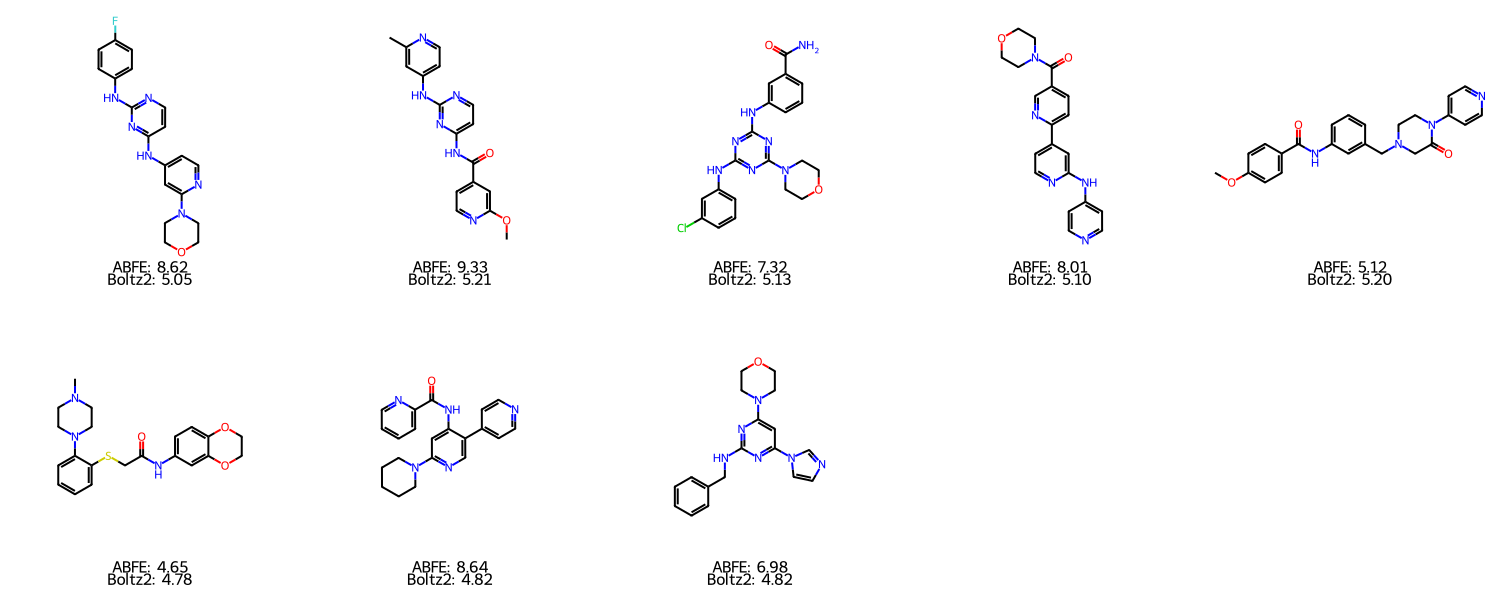}}
\caption{Selected orthosteric compounds for validation with AFEP at a Boltz2 affinity of $5 \pm 0.25$.}
\label{fig:ABFE_ortho_p5}
\end{center}
\end{figure}

\begin{figure}[ht]
\begin{center}
\centerline{\includegraphics[width=\textwidth]{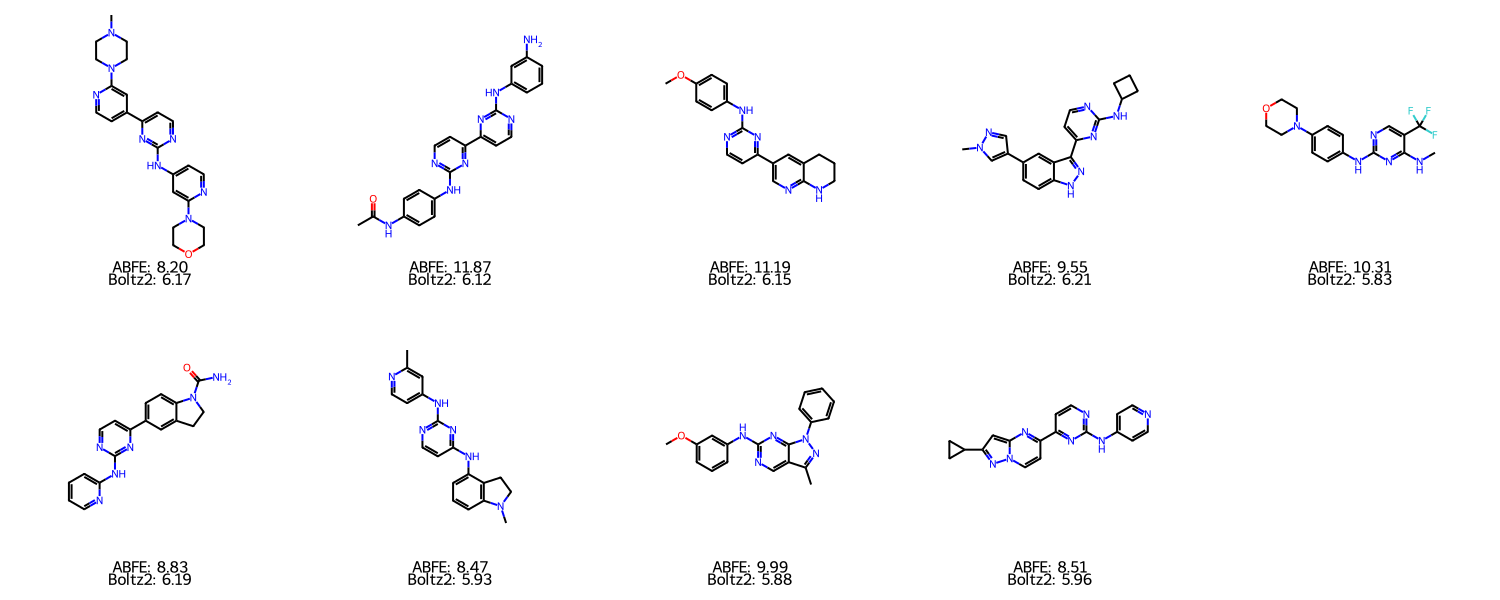}}
\caption{Selected orthosteric compounds for validation with AFEP at a Boltz2 affinity of $6 \pm 0.25$.}
\label{fig:ABFE_ortho_p6}
\end{center}
\end{figure}

\begin{figure}[ht]
\begin{center}
\centerline{\includegraphics[width=\textwidth]{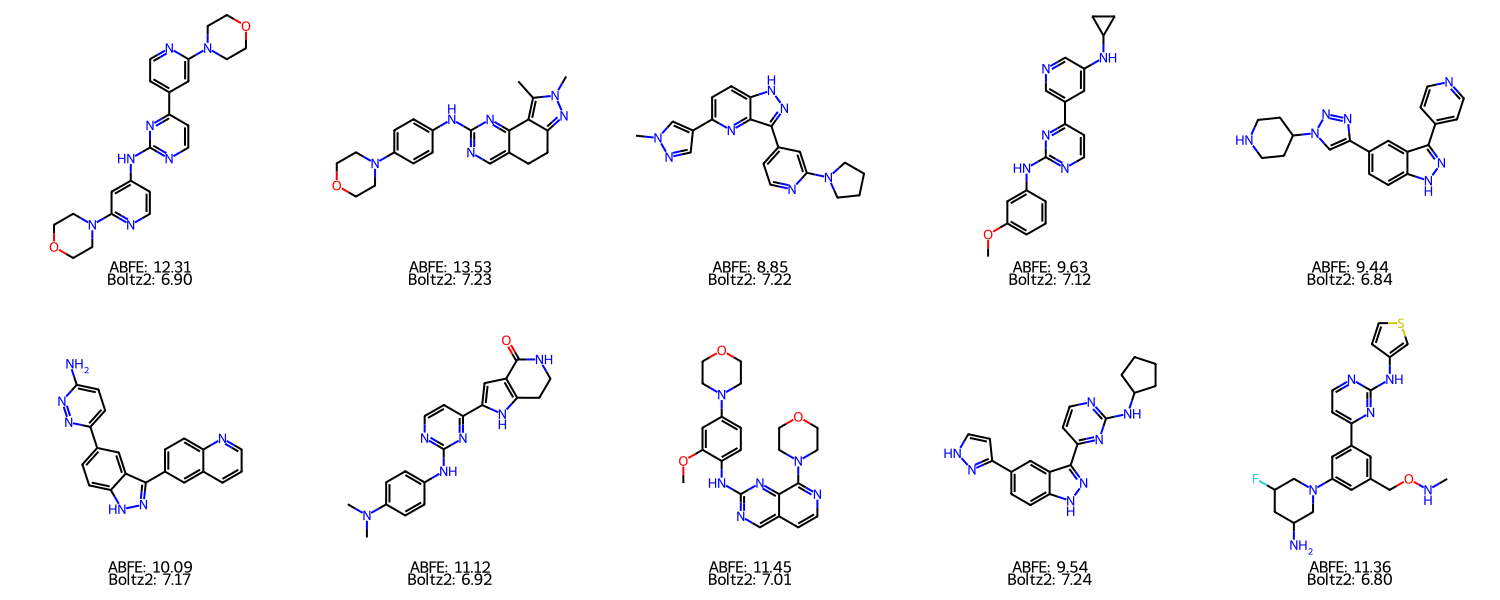}}
\caption{Selected orthosteric compounds for validation with AFEP at a Boltz2 affinity of $7 \pm 0.25$.}
\label{fig:ABFE_ortho_p7}
\end{center}
\end{figure}

\begin{figure}[ht]
\begin{center}
\centerline{\includegraphics[width=\textwidth]{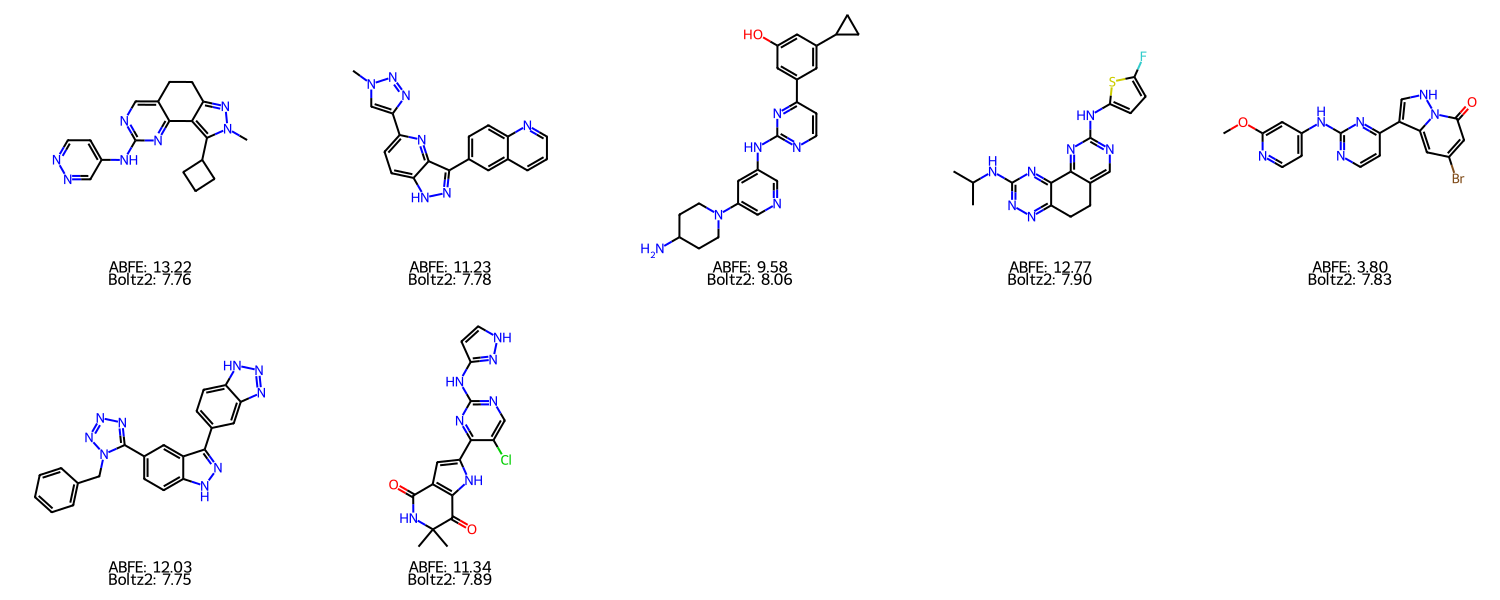}}
\caption{Selected orthosteric compounds for validation with AFEP at a Boltz2 affinity of $8 \pm 0.25$.}
\label{fig:ABFE_ortho_p8}
\end{center}
\end{figure}

\begin{figure}[ht]
\begin{center}
\centerline{\includegraphics[width=\textwidth]{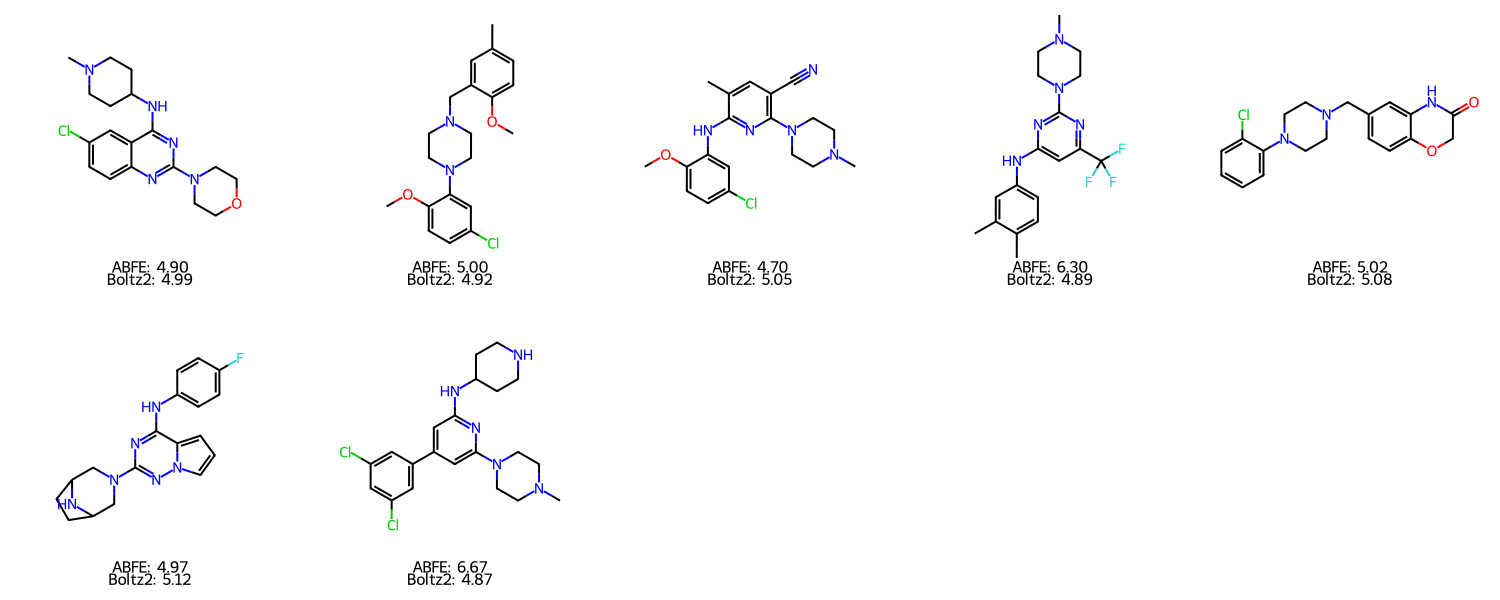}}
\caption{Selected allosteric compounds for validation with AFEP at a Boltz2 affinity of $5 \pm 0.25$.}
\label{fig:ABFE_p5}
\end{center}
\end{figure}

\begin{figure}[ht]
\begin{center}
\centerline{\includegraphics[width=\textwidth]{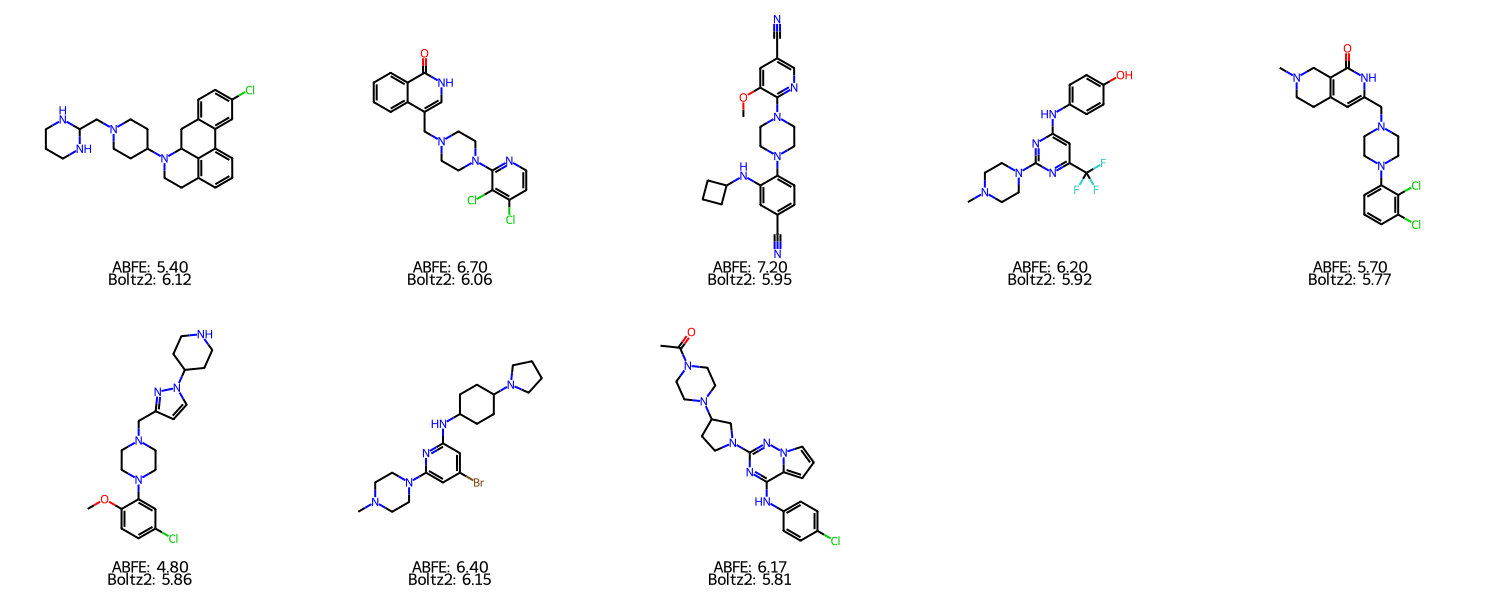}}
\caption{Selected allosteric compounds for validation with AFEP at a Boltz2 affinity of $6 \pm 0.25$.}
\label{fig:ABFE_p6}
\end{center}
\end{figure}

\begin{figure}[ht]
\begin{center}
\centerline{\includegraphics[width=\textwidth]{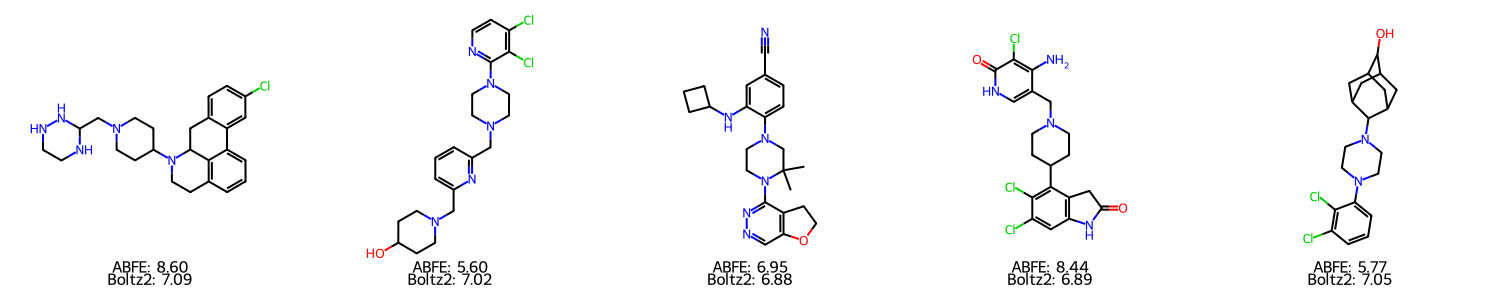}}
\caption{Selected allosteric compounds for validation with AFEP at a Boltz2 affinity of $7 \pm 0.25$.}
\label{fig:ABFE_p7}
\end{center}
\end{figure}

\begin{figure}[ht]
\begin{center}
\centerline{\includegraphics[width=\textwidth]{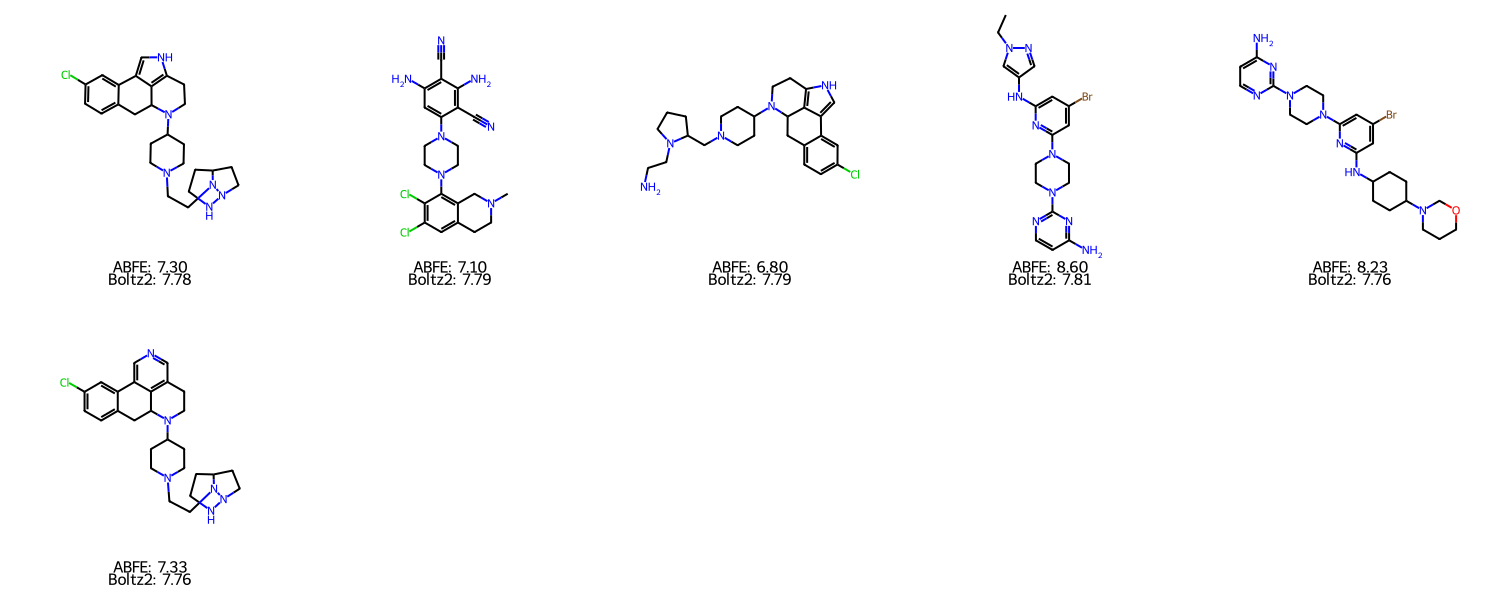}}
\caption{Selected allosteric compounds for validation with AFEP at a Boltz2 affinity of $8 \pm 0.25$.}
\label{fig:ABFE_p8}
\end{center}
\end{figure}

A random subset of up to 10 ACEGEN generated JNK3 putative orthosteric and allosteric compounds was selected within a range of $\pm$ 0.25 from four Boltz2 estimated affinity values (5, 6, 7, 8). To ensure similar physicochemical properties, molecules were sampled from a heavy atom count range between 25 and 35, and a logP range between 2 and 4 (calculated using cLogP \citep{wildman1999prediction}). Compounds that failed ABFE simulations were removed. Remaining compounds selected are shown in \autoref{fig:ABFE_ortho_p5} - \autoref{fig:ABFE_p8}.

\begin{figure*}[ht]
\begin{center}
\centerline{\includegraphics[width=\textwidth]{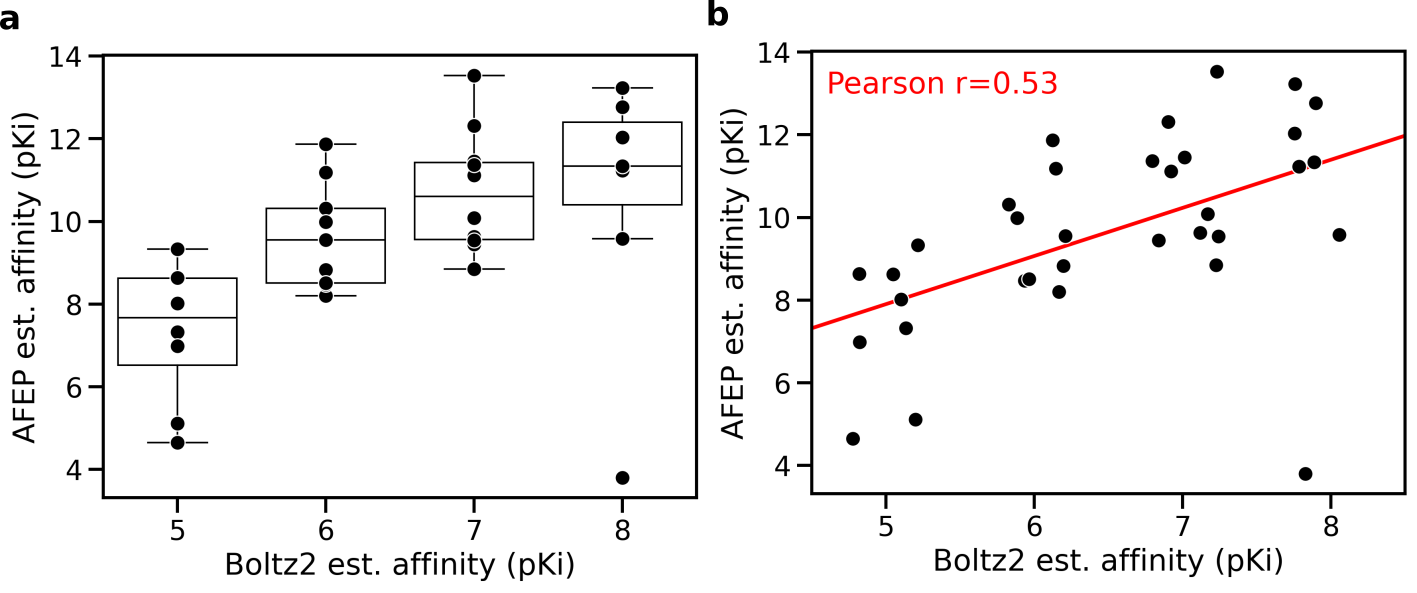}}
\caption{Correlation between Boltz2 estimated pKi and ABFE estimated pKi. (a) Binned affinities show a positive stepwise correlation, where a one-way ANOVA test reports significant variance between the means (p=0.0049). (b) The non-binned correlation between Boltz2 estimated affinity and AFEP with a reported Pearson correlation of 0.53.}
\label{fig:AFEP_validation}
\end{center}
\end{figure*}

\clearpage
\bibliography{SI_refs}